\documentclass[11pt]{article}

\PassOptionsToPackage{sort}{natbib}
\usepackage[final]{acl}

\usepackage{times}
\usepackage{latexsym}
\usepackage{caption}
\usepackage{dingbat}
\usepackage{calc}
\usepackage{enumitem}
\usepackage{booktabs}

\usepackage[most]{tcolorbox}

\newtcblisting{promptbox}[1][]{
  enhanced, breakable,
  sharp corners,
  colback=white, colframe=black,
  boxrule=0.6pt,
  colbacktitle=black, coltitle=white,
  fonttitle=\sffamily\bfseries\small,
  title={Prompt},
  left=2mm, right=2mm, top=2mm, bottom=2mm,
  listing only,
  listing options={
    basicstyle=\ttfamily\small,
    breaklines=true,
    columns=fullflexible,
    keepspaces=true,
  },
  #1
}

\usepackage[T1]{fontenc}

\usepackage[utf8]{inputenc}

\usepackage{microtype}

\usepackage{inconsolata}

\usepackage{graphicx}
\usepackage{xurl}
\usepackage{CJKutf8}

\newcounter{runin}[section]
\renewcommand{\therunin}{\thesection.\arabic{runin}}
\newcommand{\runin}[2][]{%
  \refstepcounter{runin}%
  \if\relax\detokenize{#1}\relax\else\label{#1}\fi
  \paragraph{\therunin\ \ #2}}

\newcommand{\pat}[1]{{\ttfamily\spaceskip=0.3em plus 0.1em minus 0.05em #1}}

\title{Evaluating Physically Grounded Procedural Reasoning in VLMs through Real-World Execution: Crochet as a Testbed}

\author{Maximiliano Wardle \\
  University of York / York, UK \\
  \texttt{cgf520@york.ac.uk}}

\usepackage{stfloats}

\begin{document}
\twocolumn[{%
  \renewcommand\twocolumn[1][]{#1}%
  \maketitle
  \begin{center}
    \includegraphics[width=\textwidth]{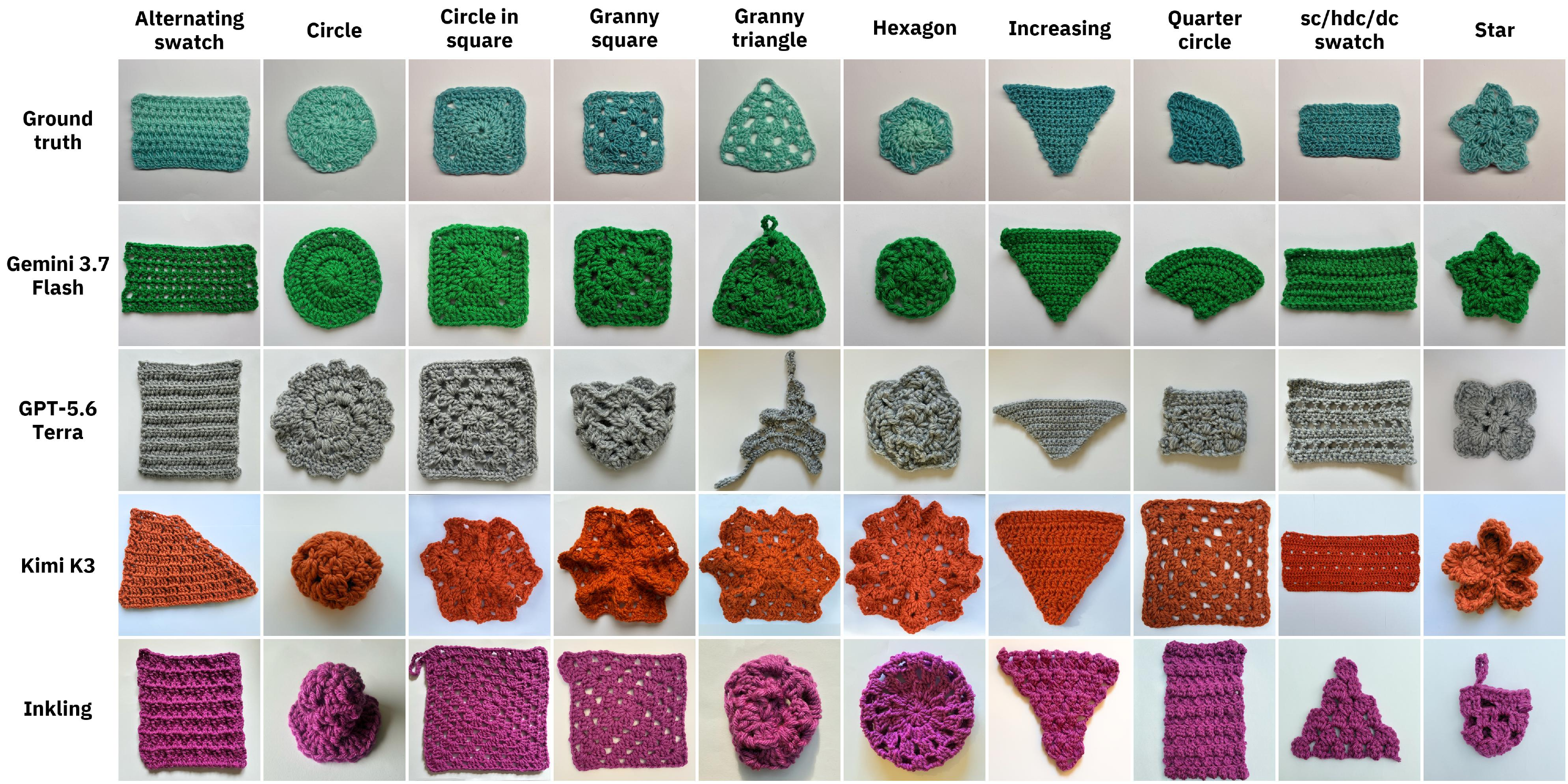}
    \captionof{figure}{The FoundationChain Testbed - Given a photo of a finished crochet object (top row), VLMs must infer how it was made and generate a coherent procedure that gives the same result. We execute 40 crochet patterns generated by 4 VLMs by hand and score the resulting objects (other rows) against a comprehensive error taxonomy.}
    \label{fig:foundationchain-grid}
  \end{center}
}]
\begin{abstract}
Vision-Language Models (VLMs) can generate procedures that look convincing on paper, but do they work in practice? In this paper, we propose crochet as an ideal testbed for physically grounded procedural reasoning in VLMs. Crochet cannot be reliably simulated or automated: the only way to know what a pattern produces is to make it by hand, yet verification is tangible and cheap, requiring nothing but a hook and a ball of yarn. We introduce FoundationChain, a deceptively simple test of procedural reasoning: given an image of a basic crochet object, can VLMs infer how it was made and generate a pattern that produces the same result? We crochet 40 patterns generated by four VLMs by hand. Even at this scale, physical execution not only reveals failures that automated proxies cannot detect, but also enables a more precise diagnosis of their causes and consequences: while models write fluent, convincing patterns, only four are free of errors, and our analysis exposes deficiencies in visual, geometric and procedural reasoning.
\end{abstract}

\section{Introduction}
\label{sec:intro}

From tying knots \citep{chen2025knot}, building LEGO \citep{legocobuilder} and assembling furniture \citep{liu2026ikea}, to origami \citep{agarwal2026aiunderstandlanguageorigami, learn2fold}, cooking \citep{simmer2026} and even lab work \citep{liu2026bioprobench}, recent work finds that while Vision-Language Models (VLMs) can generate procedures that look convincing on paper, these often fall apart in practice. These shortcomings are most clearly exposed in \emph{making} contexts, where the end goal is a physical object and success hinges not only on interpreting the object's current state, or even on inferring what the next action should be and predicting its effects, but on composing many interdependent actions into a coherent sequence that yields the desired result when executed. As we increasingly look to models to guide action in the real world, whether by controlling robots  \citep{pmlr-v205-ichter23a, internvla_a15} or directing people \citep{xu2026pro2assistcontinuousstepawareproactive, eyetoeye2026}, mapping these limits is more important than ever. Yet because of physical execution's costs (time, materials, safety, expertise) evaluation of procedural reasoning in VLMs seldom goes beyond text- or simulation-based proxies.

\begin{figure}[t]
  \centering
  \includegraphics[width=\columnwidth]{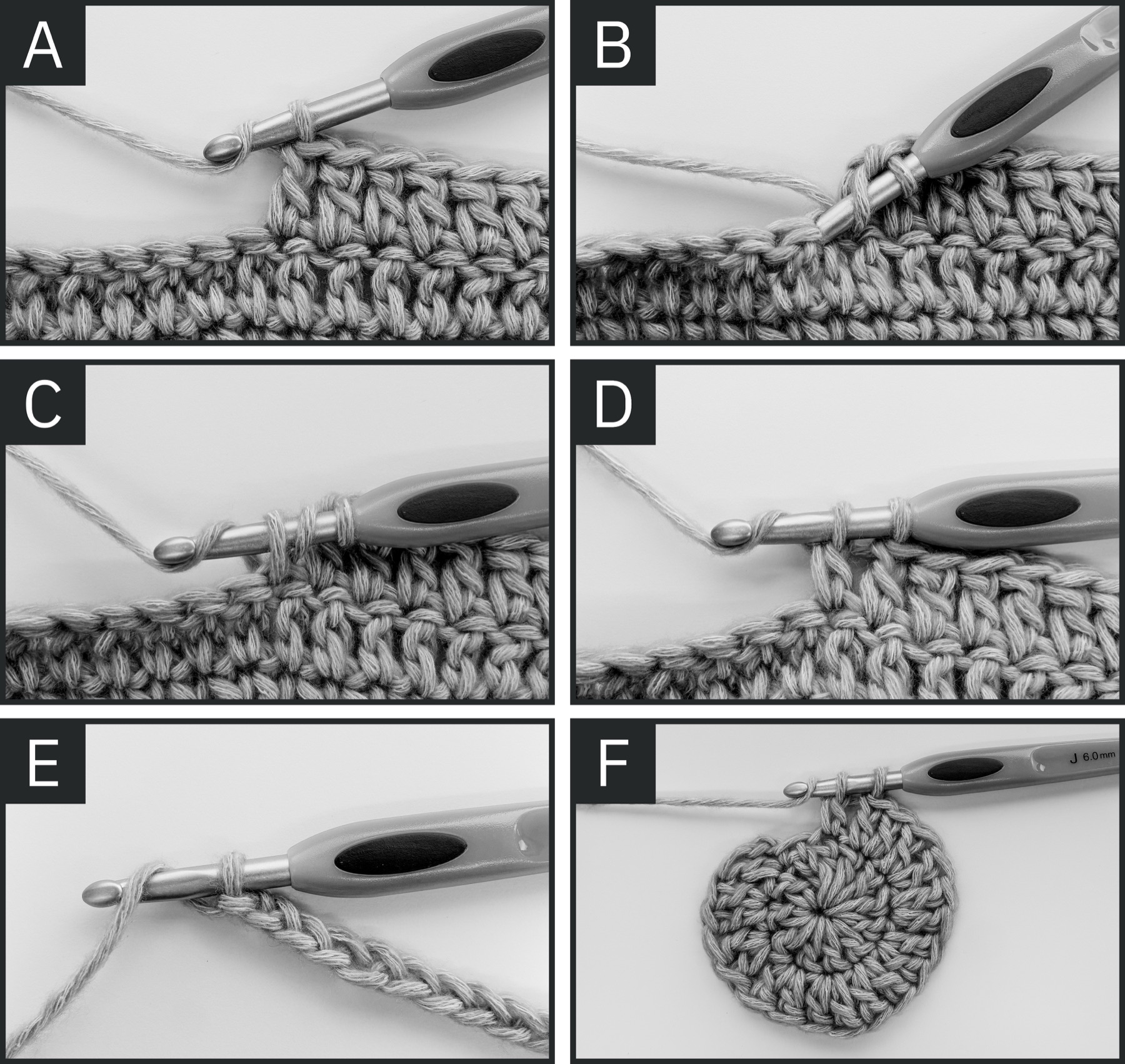}
  \caption{\textbf{A--D:} making a `double crochet' stitch. Yarn is wrapped over the hook (A), the hook is inserted into the target stitch and yarn is drawn back through (B), then through two of the loops on the hook (C), and finally through the last two loops (D), completing the stitch and returning to the state in A. \textbf{E:} making a `foundation chain', which acts as the base for the first row of stitches in a pattern. \textbf{F:} working `in the round’ to make a circle.}
  \label{fig:tiled}
\end{figure}

In this work, we propose crochet---the art of making textile objects by using a hook to draw yarn into interlocking loops, or `stitches' (Fig.~\ref{fig:tiled})---as a testbed for physically grounded procedural reasoning, one that can be easily and safely evaluated through real-world execution. Though often confused with knitting, crochet differs in ways that make it a harder challenge for VLMs. In knitting, every loop in the current row stays `live', producing a regular lattice. Crochet, by contrast, keeps just \emph{one} loop live at a time, so stitches can be worked into any available space or chained outward in any direction. This freedom makes crochet ideal for making complex 3D forms, from hats and amigurumi dolls to models of the hyperbolic plane \citep{henderson2001crocheting} and the Lorenz manifold \citep{osinga2004crocheting}. It also makes crochet very difficult to automate. Whereas knitting was mechanised in 1589 \citep{friedman2024leibniz}, and knitting machines now run programs that can even be inferred from images \citep{inverseknitting}, only a handful of prototype crochet machines have ever been built \citep[e.g.,][]{crochematic}. Crochet is also strictly sequential: each stitch depends on all those made before it, so an error made early on cannot be corrected without undoing every action since. This makes it precisely the type of task where autoregressive next-token prediction, which commits to each step before its consequences can be observed, is known to struggle \citep{ye2025beyond, pmlr-v235-bachmann24a, learn2fold}.

Crochet objects are made by people (crocheters) following instructions (patterns). Patterns are programmatic texts, executed row by row or round by round, with repeats, functions, variables and evolving states. But patterns also rely heavily on taken-for-granted conventions and the crocheter’s own embodied knowledge, and vary even between varieties of the same language.\footnote{A US ``single crochet” stitch is a UK “double crochet”; we use US terms. Conventions vary even more across languages. For example, Chinese patterns use a compact shorthand (e.g., \pat{X} for single crochet, \pat{F} for double crochet). Japanese patterns use diagrams instead of text---the symbols used are standardised in JIS L 0201 \citep{jisl0201}.} 

A pattern describes a graph of stitches (a formalisation is available in \S\ref{app:graph}), but its shape is not fixed: many shapes are consistent with the same graph. Which one an object takes is resolved only when it is physically embedded (crocheted), as tension in the yarn draws stitches of different lengths into position \citep{amigo, tassev2026layout}. In other words, the geometry a pattern produces therefore cannot be reliably predicted simply by reading it. Nor, as we show in \S\ref{sec:benchmarks}, can the automated proxies currently used to evaluate VLM-generated crochet separate a pattern that gives the desired result from one that does not. For now, the only reliable way to know is to crochet it by hand.

We present a deceptively simple test of procedural reasoning: given an image of a basic crochet object, can VLMs infer how it was made and generate a pattern that produces the same result? Our testbed, FoundationChain (\S\ref{sec:foundationchain}), uses crochet to test description, planning and pattern writing through a three-stage scaffold, and each executed pattern is scored against a taxonomy of eight error types that enables us to diagnose failures, their causes and their consequences. Crochet offers a domain where physical execution is not only necessary but cheap and safe: every result can be verified by anyone with a hook and a ball of yarn. As a first step, we crochet and annotate 40 patterns generated by four VLMs by hand, representing around 60 hours of work. Just four are error-free, and the results expose deficiencies in visual, geometric and procedural reasoning (\S\ref{sec:results}). We will release the testbed along with all generated descriptions, plans and patterns.

\section{Evaluation without Execution: The Limits of Automated Benchmarks}
\label{sec:benchmarks}

To our knowledge, only one benchmark currently tests crochet pattern generation in VLMs.\footnote{\citet{dias-karim} evaluate translation of natural-language patterns into CrochetPARADE programs, but do not test novel pattern generation.} Built on over 6,000 human-authored pattern--image pairs from the pattern website Yarnspirations \cite{spinrite2026yarnspirations}, CrochetBench \citep{crochetbench} frames its four tasks as testing whether models can go from ``describing’’ crochet (A and B) to ``doing’’ it (C and D), though no VLM-generated pattern is ever crocheted. CrochetBench argues this is unnecessary, since crochet instructions can be compiled and validated ``automatically’’ without ``costly physical execution’’ \citep[p.~13229]{crochetbench}. Using the benchmark's data and code, we test this claim on Tasks C and D and show that the automated proxies they use cannot reliably determine whether VLM-generated crochet patterns actually produce the correct object.

\paragraph{Lexical overlap rewards surface similarity, not functional equivalence.}
In Task~C, models are shown an image of a target crochet object and asked to write a pattern for it, which is scored by its lexical overlap (BLEU, ROUGE-L, ChrF) with the human-authored original. We show these metrics cannot distinguish correct patterns from incorrect ones. We score each Task~C reference pattern against another, randomly selected reference ($n = 6{,}084$ pairs, 5 seeds), yielding BLEU 4.25, ROUGE-L 20.58 and ChrF 26.43. Although every pair describes the wrong object, these scores would place second among the ten VLMs CrochetBench tests in ChrF, third in BLEU and seventh in ROUGE-L. $n$-gram analysis reveals this overlap consists mostly of generic crochet phrasing such as ``sc in each’’ and ``Fasten off’’ (\S\ref{app:ngrams}). When we replace every integer in each reference with a random value, these perturbed versions score BLEU 64.23, ROUGE-L 82.47 and ChrF 85.09 against the originals, far above every model CrochetBench evaluates. Lexical metrics' lack of discriminative power has been widely documented elsewhere, in contexts as varied as sewing \citep{geiger-etal-2025-evaluating}, code generation \citep{10.1145/3551349.3556903} and radiology \citep{ostmeier-etal-2024-green}. Code generation overcame them by moving to execution-based evaluation \citep{chen2021codex}, as we do for crochet in \S\ref{sec:foundationchain}.

\paragraph{Compiler validity tests DSL syntax, not pattern-writing ability.}
Task~D supplies a human-authored pattern for models to translate into the CrochetPARADE \citep{tassev2023crochetparade} domain-specific language (DSL), which includes a 3D renderer. It tests translation, not novel pattern generation. Outputs are scored in two stages: first, the percentage of DSL translations that compile (Valid Pattern Rate; VPR); then, the cosine similarity between DINO [CLS] \citep{caron2021emergingpropertiesselfsupervisedvision, oquab2024dinov2learningrobustvisual} embeddings of a wireframe render of each valid program and a photo of the target object.

CrochetBench acknowledges that VPR tests only syntactic validity, not semantic equivalence, but still cites low scores across the VLMs it evaluates as evidence of poor procedural reasoning \citep[p.~13235]{crochetbench}. This confounds crochet ability with unfamiliarity with CrochetPARADE, a low-resource DSL not widely adopted by crocheters. We test eight widely used VLMs on Task~D (\S\ref{app:taskd}) and find that standard remedies from domain-specific code generation lift the best model, Gemini 3.7 Flash \citep{gdm2026gemini37flash}, from just 7\% VPR under CrochetBench’s prompt to 28\% with the addition of basic syntax guidance (\citealp{wang2023grammar}; \S\ref{app:enhanced-prompt}) and as high as 64\% with one round of execution feedback \citep{gao20263dcodebench} and simple rule-based repair (\citealp[]{NEURIPS2025_APOLLO}; \S\ref{app:compilerfeedback-prompt}). Fine-tuning the weakest model overall, Qwen3.5-4B \citep{qwen2026qwen35model}, on 2,026 compiler-valid DSL programs increases its VPR from 1\% to 73\% under CrochetBench’s prompt, yet without a natural-language pattern to translate, it cannot write coherent patterns at all (\S\ref{app:finetune}).

\begin{figure}[t]
\centering
\includegraphics[width=\columnwidth]{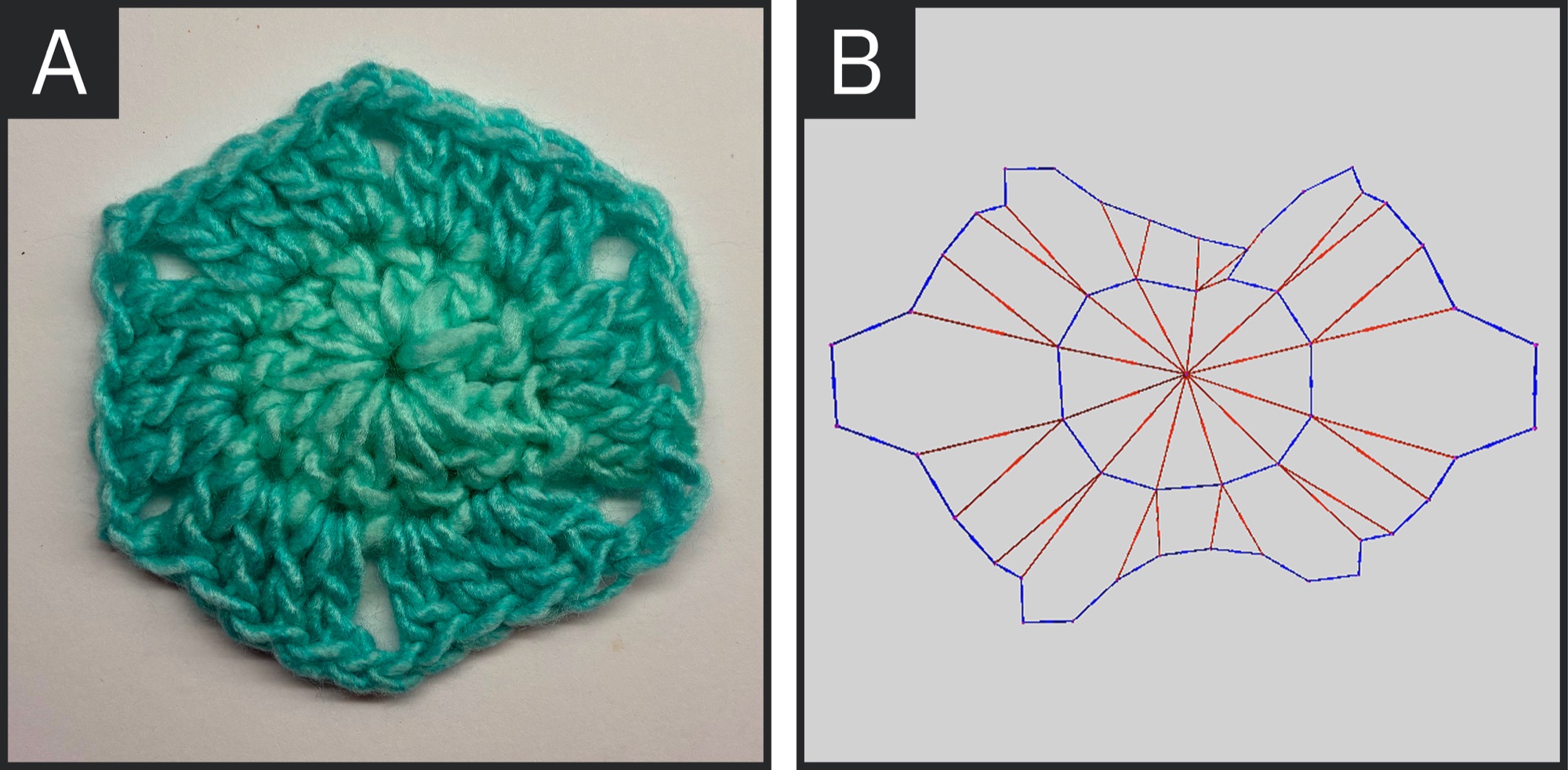}
\caption{\textbf{A: }Our own crocheted Hexagon \textbf{B:} A render of a hand-written CrochetPARADE DSL translation of the same pattern. Their cosine similarity is just 0.184.}
\label{fig:hexagon-dino}
\end{figure}

\paragraph{Render similarity cannot confirm that patterns produce the correct geometry.} CrochetBench takes low similarity between renders and photographs as evidence that translations lack ``global structural fidelity’’ \citep[p.~13237]{crochetbench}. But this holds only if wireframe renders of correct programs score higher against photographs than renders of incorrect ones. That is unlikely: DINO [CLS] embeddings capture an image’s global appearance in a single vector while largely discarding spatial structure \citep{9879898}. Moreover, CrochetPARADE’s physics solver reconstructs one of many plausible rest geometries for a given program (\citealp{tassev2026layout}; \S\ref{app:graph}), so even a render of an exact human-authored translation can look very different to the crocheted object (Fig.~\ref{fig:hexagon-dino}).

We compute DINO similarity between renders of 134 valid VLM-generated translations and photos of the 56 crochet items they target (\S\ref{app:dino}). Renders are barely more similar to photos of the item they purportedly depict than to those of unrelated items (mean 0.188 vs.\ 0.176, consistent with CrochetBench’s reported scores), and the correct photo ranks on average 26.4 of 56, versus 28.5 by chance ($d$ = 0.11, $p$ = 0.007). Renders of different items are instead far more similar to each other (mean 0.58, near the benchmark’s threshold of 0.6 for a good match), as are photos of different objects (mean 0.27). The marked difference in appearance between renders and photographs thus dwarfs any gap between correct and incorrect programs.

\section{Evaluation through Physical Execution: The FoundationChain Crochet Testbed}
\label{sec:foundationchain}
Automated benchmarks cannot reliably predict whether a VLM-generated pattern yields the correct object. Nor can they precisely diagnose where, how, or why patterns fail. Only physical execution---that is, crocheting the pattern by hand---can answer both questions. We therefore introduce FoundationChain, a testbed for foundational procedural reasoning in VLMs. Given an image of a crochet object, models must interpret what it is (\textit{perception}), infer how it might be made (\textit{planning}) and write a pattern that can reproduce it (\textit{procedure writing}). We crochet each generated pattern by hand (\textit{execution}) and score it against a taxonomy of eight error types, recording which occur and where they first appear (\textit{scoring}).

\paragraph{Targets.} Currently, FoundationChain includes ten elementary targets (Fig.~\ref{fig:foundationchain-grid}). Each is designed to probe a different aspect of crochet pattern writing, including core stitches, construction methods and shape formation (see \S\ref{app:targets}). Though simple, these targets are not `toy’ examples, but primitives from which larger projects are often assembled (e.g., joining squares into a blanket). We therefore present FoundationChain not as a leaderboard but as a minimum baseline: a model that cannot write working patterns for these targets is unlikely to succeed at more complex ones. We wrote, crocheted and photographed all targets ourselves, adapting widely-available examples. Thousands of versions of these patterns have been published online and in print, meaning models have likely encountered them before,\footnote{A search of \citeauthor{ravelry2026granny} (\citeyear{ravelry2026granny}), the largest pattern database, returns 29,005 results for `granny square’.} However, none of our images have been published until now, and so could not have been included in model training data.

\paragraph{Generation.} Generation follows a three-stage scaffold (\S\ref{app:scaffold}) which proceeds in order of increasing difficulty and enables us to test perception, planning and procedure writing together, and identify how errors propagate from one stage to the next. It also ensures failures cannot be attributed to models writing patterns cold. The stages are:

\begin{enumerate}[itemsep=0.5ex, topsep=0.5ex, parsep=0pt]
    \item \textit{Description.} The model is given an image of the target object and prompted to generate a name and a short description of no more than three sentences.
    \item \textit{Planning.} The model is shown the image again, together with its name and description, and asked to create a plan to reproduce it by answering eleven questions, such as the overall shape, the construction method used and the stitches it contains. This stage is therefore descriptive as well as prescriptive, requiring models to describe the target in greater detail \emph{and} to infer how it was made, thereby conditioning the patterns that they generate.
    \item \textit{Pattern Writing.} Using the image, description and answers to the planning questions, the model generates the final pattern for the target.
\end{enumerate}

\paragraph{Models.} We test four recent, widely-used proprietary and open-weight VLMs: Gemini 3.7 Flash \cite{gdm2026gemini37flash}, GPT-5.6 Terra \cite{openai2026gpt56}, Kimi K3 \cite{moonshot2026kimik3}  and Inkling \cite{thinkingmachines2026inkling}. These were among the largest and most capable VLMs available at time of generation, so failures cannot be attributed to model size. Each model writes a pattern for every FoundationChain target, giving 40 patterns total. To measure base capability rather than test-time compute and to improve reproducibility, we use temperature 0 and turn extended reasoning off for all models, except Gemini 3.7 Flash, which does not allow it to be disabled and which we set to 'low'  instead (full sampling settings are in \S\ref{app:sampling}). As the planning questions already externalise reasoning, this is less of a handicap than it may seem, though we note that the difference in settings nonetheless may favour Gemini.

\paragraph{Execution.} We crochet each VLM-generated pattern following it as literally as possible, using a small set of fixed rules to resolve impossible or ambiguous instructions (\S\ref{app:execution}). Whereas an invalid program simply fails to compile, a flawed pattern still produces a tangible object that can be analysed.

\paragraph{Scoring.} We score each pattern against a taxonomy of eight error types. While not exhaustive, these cover the most critical errors that determine whether a pattern produces the target object. A pattern receives one point for each type it contains, however often that type recurs, and scores range from 0 (no errors) to 8 (at least one error of every type). For each type, we also record where it first occurs: in the description, the plan, or a specific row or round of the pattern. The eight types are:

\begin{figure*}[t]
  \centering
  \includegraphics[width=\textwidth]{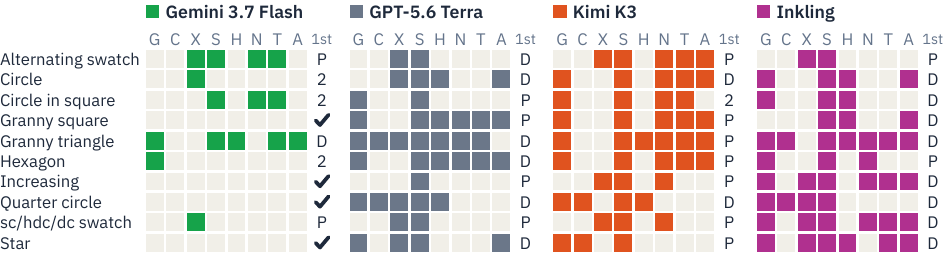}
  \caption{\textbf{Results.} Heatmaps showing error categories across each model's output patterns. Filled cells mark whether an error of a given type is present in a generated pattern (G geometry, C construction, X stitches, S size, N count, T state tracking, H hallucination, A ambiguity). The final column, \texttt{1st}, denotes the round in which the first error occurs, with D and P representing the description and planning stages. \checkmark indicates an error-free pattern.}
  \label{fig:heatmap}
\end{figure*}

\begin{itemize}[itemsep=0ex, topsep=0ex, parsep=0pt]
  \item \textit{Geometry:} the output has the wrong shape, e.g., the wrong number of sides, or a curved surface (all targets are flat).
  \item \textit{Construction:} using the wrong construction method. Working \textit{in rows}, objects grow upwards from the bottom edge. Working \textit{in the round}, objects grow outward from a central ring as for circles, polygons and most 3D shapes. Each method can be used to create some geometries but not others, so choosing incorrectly can make the target unreachable.
  \item \textit{Stitches:} selecting the wrong stitches or combinations of them. The targets use only chain (ch), single (sc), half-double (hdc), double (dc) and treble (tr) crochet and slip stitch (slst).
  \item \textit{Size:} the pattern has more or fewer rows or rounds than the target, giving an object that is too large or too small.
  \item \textit{Count:} stated counts differ from those the instructions actually produce.
  \item \textit{State tracking:} an instruction fails to account for previous steps or the object's current state.
  \item \textit{Hallucination:} the pattern adds elements not present in the target.
  \item \textit{Ambiguity:} an instruction is vague, contradictory or open to more than one reading (e.g., unclear referents, conflicting conditionals).
\end{itemize}

\section{Experiments}
\label{sec:results}
Every model generates fluent, pattern-like text for all targets (examples are provided in \S\ref{app:outputs}), yet crocheting each output by hand (around 60 hours of work) reveals only 4 of the 40 patterns are error-free (Fig.~\ref{fig:heatmap}). No VLM we test reliably produces correct patterns. All four error-free examples come from Gemini 3.7 Flash, though none is a stitch-for-stitch reproduction of the original. Gemini 3.7 Flash scores 15 points, while GPT-5.6 Terra, Kimi K3 and Inkling score 39, 43 and 44 respectively. Across all four VLMs, this gives an average of 3.53 errors per pattern. As Fig.~\ref{fig:foundationchain-grid} demonstrates, most errors are not minor deviations from the target but major failures that radically change which object emerges when the pattern is executed. 

Below, we examine where these errors occur, how they affect the crocheted objects, and what they reveal about physically grounded procedural reasoning in VLMs. We organise our findings around the three capabilities needed to write a working pattern: interpreting the target (\S\ref{par:perception}), inferring the discrete actions needed to recreate it and predicting their effects (\S\ref{par:geom}), and composing those actions into a coherent procedure that gives the correct result when executed (\S\ref{sec:snowballing}).

\subsection{Visual Reasoning} \label{par:perception} Perception is the first bottleneck. In almost half of generated outputs with errors (17 of 36), the first occurs in the description stage. In a further 15, it occurs in the plan, which itself requires models to analyse the image to infer how the target was made (see \S\ref{app:scaffold}). Because each stage is conditioned on the model's previous outputs, perceptual errors become part of the prompt used for pattern writing. In 12 of 36, models settle on an incorrect interpretation early and write patterns that, while wrong, are otherwise coherent (\S\ref{app:strongwrong}). For instance, GPT-5.6 Terra, Kimi K3 and Inkling describe the Quarter Circle as ``square'' and write their patterns accordingly, whereas Gemini 3.7 Flash correctly identifies it as a ``quadrant’’ or ``fan''. In most cases (20 of 36) however, perception is only the first link in a chain of reasoning failures that combine to produce incoherent patterns and thus, malformed objects. We find Gemini 3.7 Flash’s substantial lead over the other VLMs is due largely to more precise description: most of the errors it makes occur in pattern writing. Whether this actually reflects stronger visual reasoning, or simply better recall of similar objects in its training data is unclear. Overall, our results mirror findings that VLMs' failures in visual tasks are attributable to poor perception as much as flawed reasoning (\citealp{wang-etal-2026-reasoning-benchmark, wu2026seeing}; \citealp[cf.][]{crochetbench}).

82.5\% of all outputs (33 of 40) are the wrong size, by far the most common error the VLMs we test make. For crocheters, counting the rows or rounds a finished piece has is usually straightforward, especially for objects as small as our targets: the median is 3.5 rows or rounds, none has more than 17. But counting even small numbers of things is a known weakness of VLMs \citep{rahmanzadehgervi2024vision}. GPT-5.6 Terra, Kimi K3 and Inkling overestimate the number of rows or rounds by 3.0--5.2 on average. Gemini 3.7 Flash is more accurate (signed error $-$0.10; mean absolute error, MAE 0.50).\footnote{The magnitude of these errors seems to be partly due to the task's framing: the number of rows or rounds is just one of eleven planning questions VLMs are asked \citep[cf.][]{wang2026compose, fan2026tinted}. When we ask models \emph{only} to count how many rows or rounds each target has (see \S\ref{app:rowcount}), GPT-5.6 Terra, Kimi K3 and Inkling are more accurate (MAE 1.6--1.8), while Gemini 3.7 Flash performs, if anything, slightly worse (MAE 0.67). Changing the temperature or thinking budget does not close the gap between Gemini and the other models.} 32.5\% of outputs (13 of 40) contain hallucinated features, many of which are surprisingly consistent across models: `scalloped' borders feature in 15 of 40 patterns and 5 of 40 have a `hanging loop'.\footnote{The ``hanging loop'' appears to correspond to spaces between stitches at the top of certain targets, particularly the Granny Triangle.} Other perceptual errors are subtler and harder to detect without knowledge of crochet. Stitches determine texture as well as shape and size, yet 42.5\% of patterns use the wrong ones. GPT-5.6 Terra does so most often, in 6 of 10 patterns. Models give conflicting accounts of texture (e.g., the Alternating Swatch’s ridges run vertically, horizontally or diagonally, depending on the VLM) tend to propose far more elaborate combinations of stitches to recreate it than the targets actually use. These errors may reflect VLMs’ inability to resolve fine visual detail \citep{rahmanzadehgervi2024vision}, or a tendency to fall back on learned expectations of what images typically contain rather than what they actually show \citep{vlmsarebiased}, especially when faced with entities they cannot name \citep{shahgir2026vlms}. The fact that models tend to embellish rather than to simplify suggests the latter.

\subsection{Geometric Reasoning} \label{par:geom} If describing the target correctly poses a significant challenge for VLMs, reaching it is harder still: doing so requires selecting the right graph of stitches and sequence of actions, and modelling how yarn under tension settles into shape as the object is crocheted (see \S\ref{app:graph}). Of the 40 outputs, 23 (57.5\%) contain geometry errors. In 14 of these, this is despite the model stating the correct shape in both the description and the plan. Even for the elementary shapes we test, models make errors so fundamental that they suggest a poor grasp of basic geometry, at least as it is applied to crochet.

\begin{figure}[t]
  \centering
  \includegraphics[width=\columnwidth]{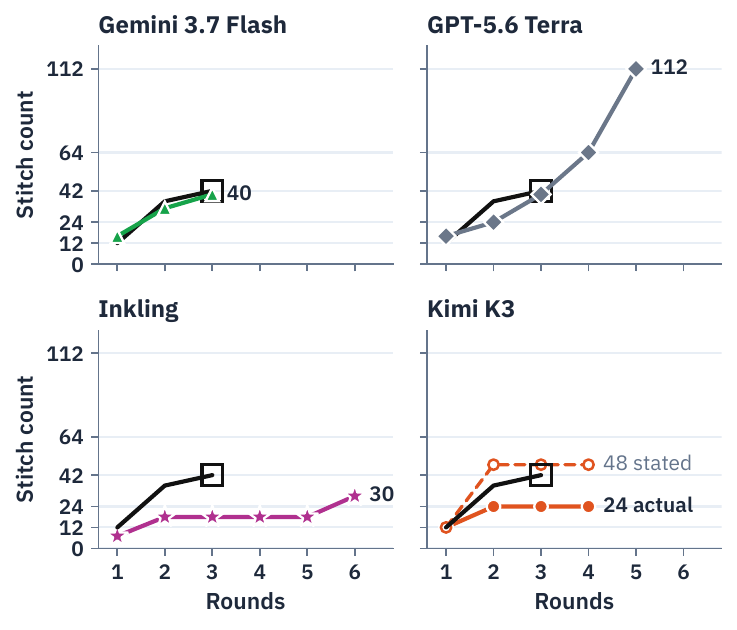}
  \caption{\textbf{Stitch counts per round for the Circle target.} Coloured lines show each model’s pattern, and the black line shows the target. Kimi K3 is doubly wrong: by skipping alternating stitches, its second round creates half as many stitches (solid line) as it claims (dashed line), though neither method would give a flat circle.}
  \label{fig:stitch-count}
\end{figure}

All of the targets are flat, yet 13 of the outputs are not. For objects made in the round, curvature is dictated by how many stitches each round adds as the object grows outwards from the centre. Increasing too slowly generates positive Gaussian curvature, producing cup- or dome-like shapes (9 patterns), while increasing too quickly creates negative curvature, causing ruffling (4 patterns, all from Kimi K3; \citealp{henderson2001crocheting}). The Circle target (Fig.~\ref{fig:stitch-count}) offers the simplest test of this principle: to remain a flat disc, each round must add roughly $2\pi H/W$ stitches, given stitch height $H$ and width $W$,  evenly distributed around the perimeter (\citealp{kekkonen2024bour}; see also \S\ref{app:graph}). Gemini 3.7 Flash’s and GPT-5.6 Terra’s patterns mostly succeed in doing so (though the latter adds hallucinated embellishments and too many rounds). Kimi K3 and Inkling, however, stop increasing after Round 2 (except for Inkling’s hallucinated ``scalloped'' edge; \S\ref{par:perception}), creating positive curvature, despite the fact that both state the same goal: a ``flat circle'' (Inkling) that ``increases evenly each round’’ (Kimi K3).

The Hexagon is barely more complex than the Circle, yet none of the generated patterns makes a recognisably six-sided shape. The target (\S\ref{app:outputs}) has just two rounds: the first is a simple circle, and the second doubles the number of stitches and adds six evenly-spaced pairs of chains to form the points. Gemini 3.7 Flash’s plan describes the correct method exactly, but its final pattern makes the sides so narrow that the yarn is under too much tension to form distinct corners, so the result is a circle. Kimi K3 and Inkling both describe the target correctly, as a ``hexagon with six distinct corners'' (Kimi K3) and ``six straight sides’’ (Inkling), but make \emph{twelve} chain spaces (a dodecagon), then add superfluous rounds that their patterns state will give six sides but do not: Kimi K3 increases too quickly, generating negative curvature, while Inkling once again does not increase at all, producing positive curvature. GPT-5.6 Terra describes the Hexagon as a ``square'' with a ``six-petal’’ centre and never reconciles the two accounts, an example of the state-tracking errors we discuss in \S\ref{sec:snowballing}.

Models state the correct goal (e.g., a flat circle) and even propose plausible strategies for reaching it (increase each round), but nonetheless generate instructions that do the opposite (no increase; Fig.~\ref{fig:stitch-count}). They reproduce the language of crochet patterns, using the right terms in the right contexts, but the link between those terms and the concrete actions they are shorthand \emph{for} seems fragile at best \citep[cf.][]{mancoridis2025potemkin, mirzadeh2025gsmsymbolic}.

A similar disconnect is evident when models must account for the ramifications \citep{handa2025actionreasoningbench} of \emph{how} the object is handled by the crocheter. The clearest illustration of this is turning: at the end of a row, the crocheter turns the piece to begin the next one, working back across the stitches just made, which are now encountered in reverse order (in \S\ref{app:graph}'s terms, turning changes only $h_t$). Although understanding this is trivial for humans, reasoning about the implicit effects of basic transformations such as rotation or reflection remains challenging for VLMs \citep{zhang2026spinbench,li2026unfolding}. 

For example, Kimi K3’s Alternating Swatch creates an asymmetrical first row, yet every subsequent instruction acts as if the order the stitches are in is invariant to turning. This is especially ironic given that the pattern specifies a ``right'' and ``wrong’’ side, which would imply the two are different. As the pattern makes heavy use of relative instructions such as ``each of the next'' and ``skip’’, executing it faithfully means ignoring several stitches each row, so every new row ends up shorter than the last, despite the pattern’s claim that the ``count remains the same throughout’’. Inkling and GPT-5.6 Terra make similar errors with the Granny Triangle: neither accounts for the effect turning has on stitch order, so neither achieves its own stated goal of making a symmetrical, three-sided shape. In each case, models put instructions to turn where pattern-writing convention dictates, but fail to account for their ramifications in the real world.

\subsection{Procedural Reasoning} \label{sec:snowballing} Composing a coherent procedure depends on state tracking: each action must follow from the state left by the one before and produce the outcome the next one relies on. \citet{petty2026finegrainedhierarchybenchmarkstate} describe four levels of state tracking: r-trivial (variables are immutable), aperiodic (variables are mutable), soluble (updates to variables can be self-referential) and general finite (variables can take on the values of other variables). \citeauthor{petty2026finegrainedhierarchybenchmarkstate} find frontier models’ accuracy declines sharply from the soluble level onwards as sequences grow longer. Crochet requires at least soluble state tracking, and many common pattern-writing conventions arguably necessitate general finite state tracking (e.g., a `turning chain' that `counts as’ the first double crochet of a new row, a space between stitches that is later referred to as a `corner’). The models we test frequently lose track of state within the first few rounds of the patterns they generate. These state tracking errors---together with closely related count and ambiguity errors---have the most catastrophic effects of the failures we measure and are the hardest to detect as they remain hidden in text and surface only when a pattern is physically executed \citep[cf.][]{simmer2026}.

State tracking errors usually manifest in a consistent way. A model writes an incorrect instruction, but ends the row by generating a plausible-seeming summary of what that row might produce were it correct anyway---usually a stitch count. In the next row, the model assumes this summary is right, and as crochet patterns rely heavily on relative instructions (e.g., `in each', `in next') and loops (`repeat from * around'), it can generate a boilerplate repeating sequence (e.g., ``*Work 3 dc in next ch-1 sp, ch 2; repeat from * around’’) along with a summary of the outcome it supposedly produces, without ever verifying the result. This cycle continues until the pattern ends, as each error conditions both the next instruction the model generates and the geometry to which the instruction is applied (cf. \citealp{snowballing}).

In four patterns, Kimi K3 begins with a circular centre which it fails to transform into the stated shape. Instead, the repeating sequences it writes create too many stitches, and the objects become $n$-gons with negative curvature. The opposite occurs with GPT-5.6 Terra’s Hexagon and Granny Square: early rounds leave too few spaces for the later rounds that depend on them. Similarly, the ``peaks'' and ``valleys’’ Inkling describes in its pattern for the Star never materialise, because its instructions create a curved object without distinct points. Even Gemini 3.7 Flash makes basic state tracking and arithmetic errors in three of its patterns, though the consequences are less extreme: the final round of its Granny Triangle fails to create a distinct third corner, causing pinching and thus positive curvature.

The relative instructions and loop constructs implicated in many of these errors are themselves among the most frequent $n$-grams in human-written patterns (\S\ref{app:ngrams}, Tab.~\ref{tab:shared-ngrams}). That models write `repeat from *’ even when no asterisk precedes it (Inkling in 5 outputs, GPT-5.6 Terra in 2) is especially telling: it suggests that autoregressive next-token prediction favours high-probability continuations at the expense of specificity and coherence crochet demands (cf.\ \citealp{mccoy2024embers}).

\section{Conclusion}
We introduce FoundationChain, a crochet-based testbed for physically grounded procedural reasoning in Vision-Language Models, comprising ten targets, a three-stage describe–plan–write scaffold and an eight-category error taxonomy. We test perception, planning and pattern writing in four VLMs, and crochet and annotate all 40 of their outputs by hand. Even at this scale, physical execution not only reveals failures that automated proxies cannot detect, but also enables a more precise diagnosis of their causes and consequences: while models write fluent, convincing patterns, only four of 40 are free of errors, and our analysis reveals key deficiencies in visual, geometric and procedural reasoning. Even the elementary shapes we test  pose significant challenges: models struggle to describe the targets correctly, infer the actions needed to recreate them, predict the real-world consequences of simple actions, and compose many interdependent actions into a coherent sequence that yields the desired result when executed. As the field increasingly relies on automated benchmarks to map the limits of VLMs, we demonstrate the promise of crochet as a domain for testing these capabilities in the real world: one that demands physical execution, but where verification is open to anyone with a hook and a ball of yarn.

\section*{Limitations}
For now, FoundationChain contains just ten elementary targets, and we only sample one pattern per target from each of four models. This is small by the standards of automated benchmarks, but reflects the cost of physical execution: each of the 40 patterns took between 30 and 90 minutes to crochet and annotate. We also only sample models at temperature 0 and with extended reasoning disabled or set to the lowest level and so cannot rule out that other settings may improve performance, though results in other contexts that combine visual, geometric and procedural reasoning suggest the gains may be limited \citep[cf.][]{gao20263dcodebench, zhang2026benchcadcomprehensiveindustrystandardbenchmark, wu2026seeing}. 

All patterns were executed and scored by a single author, who knew the ground-truth patterns and which model produced each output, which may bias execution and scoring. However, most failures were too severe to be rescued by any favourable reading of ambiguous or incoherent instructions. Any lapse would more likely lead us to correct errors than introduce new ones. This, combined with the fact that our taxonomy records whether each type of error occurs at all, not each individual occurrence and its magnitude, means the error rates we report here are, if anything, conservative. In future, blind scoring by multiple independent crocheters would further reduce this risk.

We aim to expand our experiments to encompass a greater number of targets, models and sampling settings used, and to distribute execution and scoring across a larger group of crocheters, so as to enable more comprehensive evaluation. Therefore, for now, we offer FoundationChain as a proof of method rather than a leaderboard.

\section*{Ethical Considerations}

\paragraph{Data.} We wrote all FoundationChain targets and crocheted and photographed them ourselves together with the model outputs. We will release all code, targets, prompts, generated patterns, error annotations and photographs of every crocheted output. We use the datasets, models and code the CrochetBench authors provide via their GitHub \cite{li2026crochetbench_code} and HuggingFace \cite{li2026qwen2vl_crochetbench} which they distribute under open source licenses. We use patterns and images from Yarnspirations \cite{spinrite2026yarnspirations} to enable our analysis of CrochetBench under the same non-commercial research-only terms granted to the authors \citep{crochetbench} for use, and do not redistribute them with our own work. 

\paragraph{Impact on the crochet community.} As crocheters ourselves, we recognise that research into pattern generation may be concerning for many in the crochet community, especially pattern designers, translators and others who earn a living from crochet. Tools that can turn a photo of a finished object into a pattern risk enabling designs to be copied without consent or payment. At the same time, in recent years, online marketplaces such as Etsy have been flooded with poor-quality AI-generated patterns, many of which go undetected until buyers attempt to crochet them \citep{cbc_ai_fibre_arts, abc2026etsy}. We do not set out to improve these capabilities, but to give an evidence-based account of where they currently stand, and argue that human skill and judgement are essential to doing so.

\bibliography{custom}

@inproceedings{crochetbench,
  title     = {{CrochetBench}: Can Vision-Language Models Move from Describing to Doing in Crochet Domain?},
  author    = {Li, Peiyu and Huang, Xiaobao and Hua, Ting and Chawla, Nitesh V.},
  editor    = {Liakata, Maria and Moreira, Viviane P. and Zhang, Jiajun and Jurgens, David},
  booktitle = {Proceedings of the 64th Annual Meeting of the {A}ssociation for {C}omputational {L}inguistics (Volume 1: Long Papers)},
  month     = jul,
  year      = {2026},
  address   = {San Diego, California, United States},
  publisher = {Association for Computational Linguistics},
  pages     = {13229--13251},
  doi       = {10.18653/v1/2026.acl-long.604},
  url       = {https://aclanthology.org/2026.acl-long.604/}
}

@misc{li2026crochetbench_code,
  author       = {Li, Peiyu},
  title        = {{CrochetBench}: Code and Data},
  year         = {2026},
  month        = sep,
  howpublished = {\url{https://github.com/Peiyu-Georgia-Li/crochetBench}},
  note         = {GitHub repository. Commit 4f834d5, September 9, 2026}
}

@misc{li2026qwen2vl_crochetbench,
  author       = {Li, Peiyu},
  title        = {{Qwen2-VL-7B} Finetuned on {CrochetBench} {Task C}},
  year         = {2026},
  month        = jun,
  publisher    = {Hugging Face},
  howpublished = {\url{https://huggingface.co/peiyugeorgia/qwen2vl-crochetbench-task-c-finetuned}},
  note         = {Fine-tuned from Qwen/Qwen2-VL-7B-Instruct. Initial commit June 27, 2026}
}

@article{dias-karim,
  title   = {Translation of User Crochet Patterns to {CrochetPARADE} Syntax Using Large Language Models},
  author  = {Dias, Rachael and Karim, Kayvan},
  journal = {Proceedings of the AAAI Symposium Series},
  volume  = {6},
  number  = {1},
  pages   = {200--208},
  year    = {2025},
  doi     = {10.1609/aaaiss.v6i1.36054}
}

@misc{tassev2023crochetparade,
  author       = {Tassev, Svetlin},
  title        = {{CrochetPARADE}: Crochet {PA}ttern {R}enderer, {A}nalyzer, and {DE}bugger},
  year         = {2023},
  howpublished = {\url{https://codeberg.org/crochetparade/CrochetPARADE}},
  note         = {Software. Accessed October 2026}
}

@misc{tassev2026layout,
  author       = {Tassev, Svetlin},
  title        = {{CrochetPARADE} Node Layout Documentation (v2)},
  year         = {2026},
  howpublished = {\url{https://codeberg.org/crochetparade/CrochetPARADE/src/branch/main/docs/node_layout_documentation_v2.md}},
  note         = {Accessed October 2026}
}

@inproceedings{amigo,
  author    = {Edelstein, Michal and Peleg, Hila and Itzhaky, Shachar and Ben-Chen, Mirela},
  title     = {{AmiGo}: Computational Design of Amigurumi Crochet Patterns},
  booktitle = {Proceedings of the 7th Annual ACM Symposium on Computational Fabrication},
  series    = {SCF '22},
  articleno = {5},
  year      = {2022},
  publisher = {Association for Computing Machinery},
  address   = {New York, NY, USA},
  doi       = {10.1145/3559400.3562005},
  url       = {https://doi.org/10.1145/3559400.3562005}
}

@inproceedings{representingstitch,
  author    = {Guo, Runbo and Lin, Jenny and Narayanan, Vidya and McCann, James},
  title     = {Representing Crochet with Stitch Meshes},
  booktitle = {Proceedings of the 5th Annual ACM Symposium on Computational Fabrication},
  series    = {SCF '20},
  articleno = {4},
  year      = {2020},
  publisher = {Association for Computing Machinery},
  address   = {New York, NY, USA},
  doi       = {10.1145/3424630.3425409},
  url       = {https://doi.org/10.1145/3424630.3425409}
}

@inproceedings{10.2312:exw.20251057,
  title     = {Modeling Crochet Patterns with a Force-directed Graph Layout},
  author    = {Greer, {\'E}mile and Mould, David},
  editor    = {Catalano, Chiara Eva and Parakkat, Amal Dev},
  booktitle = {ACM/EG Expressive Symposium -- WICED: Eurographics Workshop on Intelligent Cinematography and Editing},
  year      = {2025},
  publisher = {The Eurographics Association},
  doi       = {10.2312/exw.20251057}
}

@article{storck2022topology,
  author  = {Storck, J. L. and Gerber, D. and Steenbock, L. and Kyosev, Y.},
  title   = {Topology Based Modelling of Crochet Structures},
  journal = {Journal of Industrial Textiles},
  year    = {2022},
  volume  = {52},
  doi     = {10.1177/15280837221139250}
}

@inproceedings{10.2312:pg.20261024,
  title     = {{CT2Yarn}: Yarn-Level Reconstruction of Crochet from Computed Tomography},
  author    = {Luo, Chang and Umetani, Nobuyuki},
  editor    = {He, Ying and Thuerey, Nils and Liu, Lingjie},
  booktitle = {Pacific Graphics 2026 -- Conference Papers and Posters},
  year      = {2026},
  publisher = {The Eurographics Association},
  doi       = {10.2312/pg.20261024}
}

@inproceedings{10.1145/3563835.3567657,
  author    = {Seitz, Klara and Rein, Patrick and Lincke, Jens and Hirschfeld, Robert},
  title     = {Digital Crochet: Toward a Visual Language for Pattern Description},
  booktitle = {Proceedings of the 2022 ACM SIGPLAN International Symposium on New Ideas, New Paradigms, and Reflections on Programming and Software},
  series    = {Onward! 2022},
  pages     = {48--62},
  year      = {2022},
  publisher = {Association for Computing Machinery},
  address   = {New York, NY, USA},
  doi       = {10.1145/3563835.3567657},
  url       = {https://doi.org/10.1145/3563835.3567657}
}

@inproceedings{crochematic,
  author    = {Perry, Gabriella and Garc{\'\i}a del Castillo y L{\'o}pez, Jose Luis and Melenbrink, Nathan},
  title     = {{Croche-Matic}: A Robot for Crocheting {3D} Cylindrical Geometry},
  booktitle = {2023 IEEE International Conference on Robotics and Automation (ICRA)},
  year      = {2023},
  pages     = {7440--7446},
  doi       = {10.1109/ICRA48891.2023.10160345}
}

@inproceedings{inverseknitting,
  title     = {Neural Inverse Knitting: From Images to Manufacturing Instructions},
  author    = {Kaspar, Alexandre and Oh, Tae-Hyun and Makatura, Liane and Kellnhofer, Petr and Matusik, Wojciech},
  booktitle = {Proceedings of the 36th International Conference on Machine Learning},
  series    = {Proceedings of Machine Learning Research},
  volume    = {97},
  pages     = {3272--3281},
  year      = {2019},
  editor    = {Chaudhuri, Kamalika and Salakhutdinov, Ruslan},
  publisher = {PMLR},
  url       = {https://proceedings.mlr.press/v97/kaspar19a.html}
}

@article{osinga2004crocheting,
  author  = {Osinga, Hinke M. and Krauskopf, Bernd},
  title   = {Crocheting the {Lorenz} Manifold},
  journal = {The Mathematical Intelligencer},
  year    = {2004},
  volume  = {26},
  number  = {4},
  pages   = {25--37},
  doi     = {10.1007/bf02985416}
}

@article{henderson2001crocheting,
  author  = {Henderson, David W. and Taimi{\c{n}}a, Daina},
  title   = {Crocheting the Hyperbolic Plane},
  journal = {The Mathematical Intelligencer},
  year    = {2001},
  volume  = {23},
  number  = {2},
  pages   = {17--28}
}

@article{kekkonen2024bour,
  author  = {Kekkonen, Hanne},
  title   = {Crocheting {Bour's} Minimal Surfaces},
  journal = {The Mathematical Intelligencer},
  volume  = {46},
  number  = {4},
  pages   = {306--312},
  year    = {2024},
  doi     = {10.1007/s00283-023-10314-1}
}

@article{friedman2024leibniz,
  author  = {Friedman, M.},
  title   = {{Leibniz} and the Stocking Frame: Computation, Weaving and Knitting in the 17th Century},
  journal = {Minds and Machines},
  year    = {2024},
  volume  = {34},
  number  = {Suppl 1},
  pages   = {11--28},
  doi     = {10.1007/s11023-023-09623-3}
}

@misc{jisl0201,
  author    = {{Japanese Industrial Standards Committee}},
  title     = {{JIS L 0201:1995} [Stitch Symbols]},
  year      = {1995},
  publisher = {Japanese Standards Association},
  address   = {Tokyo}
}

@misc{ravelry2026granny,
  author       = {{Ravelry}},
  title        = {Pattern search: ``granny square'' (crochet)},
  year         = {2026},
  howpublished = {\url{https://www.ravelry.com/patterns/search?craft=crochet&query=granny+square}},
  note         = {29,005 results. Accessed October 12, 2026}
}

@misc{spinrite2026yarnspirations,
  author       = {{Spinrite Inc.}},
  title        = {{Yarnspirations}},
  year         = {2026},
  howpublished = {\url{https://www.yarnspirations.com}},
  note         = {Accessed November 4, 2026}
}

@misc{cbc_ai_fibre_arts,
  author       = {{CBC News}},
  title        = {Is {AI} Ruining Fibre Arts like Knitting and Crochet?},
  year         = {2026},
  month        = sep,
  howpublished = {\url{https://www.cbc.ca/player/play/video/9.7351206}},
  note         = {News video, 2:42. Accessed October 9, 2026}
}

@misc{abc2026etsy,
  author       = {Ten Broeke, Ebony},
  title        = {Once a Home for Handmade, Artists Say {Etsy} Is Now Overrun with {AI} and Copied Works},
  year         = {2026},
  month        = oct,
  howpublished = {\url{https://www.abc.net.au/news/2026-10-09/ai-copied-products-found-on-makers-platform-etsy/107191232}},
  note         = {ABC News}
}

@inproceedings{chen2025knot,
  author    = {Chen, Zizhao and Artzi, Yoav},
  title     = {Knot So Simple: A Minimalistic Environment for Spatial Reasoning},
  booktitle = {Advances in Neural Information Processing Systems 38 (NeurIPS 2025)},
  year      = {2025},
  note      = {Poster},
  url       = {https://neurips.cc/virtual/2025/loc/san-diego/poster/121700}
}

@inproceedings{legocobuilder,
  author    = {Huang, Haochen and Su, Yue and Sun, Xin and Ahsan, Moonisa and Aliannejadi, Mohammad and Viola, Irene and Ren, Zhaochun and Yu, Chuang and Lisowska, Aneta and Belopolsky, Artem and Hindriks, Koen and Cesar, Pablo and Wang, Junxiao and Pei, Jiahuan},
  title     = {{LEGO} Co-builder: Exploring Fine-Grained Vision-Language Modeling for Multimodal Assembly Assistants},
  booktitle = {Proceedings of the 28th International Conference on Multimodal Interaction},
  series    = {ICMI '26},
  pages     = {87--98},
  year      = {2026},
  publisher = {Association for Computing Machinery},
  address   = {New York, NY, USA},
  doi       = {10.1145/3776574.3831119},
  url       = {https://doi.org/10.1145/3776574.3831119}
}

@misc{liu2026ikea,
  title         = {Benchmarking and Mechanistic Analysis of Vision-Language Models for Cross-Depiction Assembly Instruction Alignment},
  author        = {Liu, Zhuchenyang and Zhang, Yao and Xiao, Yu},
  year          = {2026},
  eprint        = {2604.00913},
  archivePrefix = {arXiv},
  primaryClass  = {cs.CV},
  url           = {https://arxiv.org/abs/2604.00913}
}

@misc{agarwal2026aiunderstandlanguageorigami,
  title         = {Can {AI} Understand the Language of Origami?},
  author        = {Agarwal, Naaisha and Wu, Yihan and Guan, Xin and Garg, Ayaan and Hu, Yikuan and Li, Mohan and Collura, Vincenzo and Dai, Wang-Zhou and Ding, Yao-Xiang and Sansone, Emanuele},
  year          = {2026},
  eprint        = {2603.13856},
  archivePrefix = {arXiv},
  primaryClass  = {cs.LG},
  url           = {https://arxiv.org/abs/2603.13856}
}

@inproceedings{learn2fold,
  author    = {Huang, Yanjia and Chen, Yunuo and Jiang, Ying and Han, Jinru and Tu, Zhengzhong and Yang, Yin and Jiang, Chenfanfu},
  title     = {{Learn2Fold}: Structured Origami Generation with World Model Planning},
  booktitle = {Computer Vision -- ECCV 2026},
  editor    = {Favaro, Paolo and Kukelova, Zuzana and Maki, Atsuto and Rohrbach, Anna and Schindler, Konrad and Tombari, Federico},
  publisher = {Springer Nature Switzerland},
  address   = {Cham},
  year      = {2026},
  pages     = {40--56},
  doi       = {10.1007/978-3-032-37350-2_3}
}

@inproceedings{simmer2026,
  title     = {{SIMMER}: Benchmarking Latent Failures in {LLM} Executable Planning with a World Model},
  author    = {Lu, Xiaoxin and Zhang, Ranran Haoran and Zhang, Rui},
  booktitle = {Conference on Language Modeling (COLM)},
  year      = {2026}
}

@inproceedings{liu2026bioprobench,
  author    = {Liu, Yuyang and Lv, Liuzhenghao and Zhang, Xiancheng and Wang, Jingya and Yuan, Li and Tian, Yonghong},
  title     = {{BioProBench}: A Corpus and Benchmark for Biological Protocol Reasoning in Autonomous Science},
  booktitle = {Proceedings of the 43rd International Conference on Machine Learning},
  year      = {2026},
  note      = {ICML 2026 Poster},
  url       = {https://openreview.net/forum?id=8EZamwNqld}
}

@inproceedings{geiger-etal-2025-evaluating,
  title     = {Evaluating Spatiotemporal Consistency in Automatically Generated Sewing Instructions},
  author    = {Geiger, Luisa and Hartmann, Mareike and Sullivan, Michael and Koller, Alexander},
  editor    = {Christodoulopoulos, Christos and Chakraborty, Tanmoy and Rose, Carolyn and Peng, Violet},
  booktitle = {Proceedings of the 2025 Conference on Empirical Methods in Natural Language Processing},
  month     = nov,
  year      = {2025},
  address   = {Suzhou, China},
  publisher = {Association for Computational Linguistics},
  pages     = {18508--18525},
  doi       = {10.18653/v1/2025.emnlp-main.934},
  url       = {https://aclanthology.org/2025.emnlp-main.934/}
}

@misc{gao20263dcodebench,
  title         = {{3DCodeBench}: Benchmarking Agentic Procedural {3D} Modeling via Code},
  author        = {Gao, Yipeng and Shu, Lei and Ye, Genzhi and Xiong, Xi and Makadia, Ameesh and Guo, Meiqi and Itti, Laurent and Chen, Jindong},
  year          = {2026},
  eprint        = {2606.01057},
  archivePrefix = {arXiv},
  primaryClass  = {cs.CV},
  url           = {https://arxiv.org/abs/2606.01057}
}

@misc{zhang2026benchcadcomprehensiveindustrystandardbenchmark,
  title         = {{BenchCAD}: A Comprehensive, Industry-Standard Benchmark for Programmatic {CAD}},
  author        = {Zhang, Haozhe and Liu, Kaichen and Chen, Miaomiao and Li, Lei and Yang, Shaojie and Peng, Cheng and Chen, Hanjie},
  year          = {2026},
  eprint        = {2605.10865},
  archivePrefix = {arXiv},
  primaryClass  = {cs.AI},
  url           = {https://arxiv.org/abs/2605.10865}
}

@inproceedings{pmlr-v205-ichter23a,
  title     = {Do As {I} Can, Not As {I} Say: Grounding Language in Robotic Affordances},
  author    = {Ahn, Michael and Brohan, Anthony and Brown, Noah and Chebotar, Yevgen and Cortes, Omar and David, Byron and Finn, Chelsea and Fu, Chuyuan and Gopalakrishnan, Keerthana and Hausman, Karol and Herzog, Alex and Ho, Daniel and Hsu, Jasmine and Ibarz, Julian and Ichter, Brian and Irpan, Alex and Jang, Eric and Jauregui Ruano, Rosario and Jeffrey, Kyle and Jesmonth, Sally and Joshi, Nikhil J. and Julian, Ryan and Kalashnikov, Dmitry and Kuang, Yuheng and Lee, Kuang-Huei and Levine, Sergey and Lu, Yao and Luu, Linda and Parada, Carolina and Pastor, Peter and Quiambao, Jornell and Rao, Kanishka and Rettinghouse, Jarek and Reyes, Diego and Sermanet, Pierre and Sievers, Nicolas and Tan, Clayton and Toshev, Alexander and Vanhoucke, Vincent and Xia, Fei and Xiao, Ted and Xu, Peng and Xu, Sichun and Yan, Mengyuan and Zeng, Andy},
  booktitle = {Proceedings of The 6th Conference on Robot Learning},
  series    = {Proceedings of Machine Learning Research},
  volume    = {205},
  pages     = {287--318},
  year      = {2023},
  editor    = {Liu, Karen and Kulic, Dana and Ichnowski, Jeff},
  publisher = {PMLR},
  url       = {https://proceedings.mlr.press/v205/ichter23a.html}
}

@misc{internvla_a15,
  title        = {{InternVLA-A1.5}: Unifying Understanding, Latent Foresight, and Action for Compositional Generalization},
  author       = {{InternVLA-A1.5 team}},
  year         = {2026},
  howpublished = {\url{https://github.com/InternRobotics/InternVLA-A1.5}}
}

@misc{xu2026pro2assistcontinuousstepawareproactive,
  title         = {{Pro$^2$Assist}: Continuous Step-aware Proactive Assistance with Multi-modal Egocentric Perception for Long-horizon Procedural Tasks},
  author        = {Xu, Lilin and Yang, Bufang and Jiang, Siyang and Liu, Kaiwei and Hou, Kaiyuan and Fan, Yuang and Chen, Hongkai and Yan, Zhenyu and Jiang, Xiaofan},
  year          = {2026},
  eprint        = {2605.04227},
  archivePrefix = {arXiv},
  primaryClass  = {cs.AI},
  url           = {https://arxiv.org/abs/2605.04227}
}

@inproceedings{eyetoeye2026,
  author    = {Teng, Zhuyu and Chen, Pei and Cai, Yichen and Lu, Ruoqing and Jiang, Zhaoqu and Li, Jiayang and You, Weitao and Sun, Lingyun},
  title     = {Seeing Eye to Eye: Enabling Cognitive Alignment through Shared First-Person Perspective in Human--{AI} Collaboration},
  booktitle = {Proceedings of the 2026 CHI Conference on Human Factors in Computing Systems},
  series    = {CHI '26},
  articleno = {19},
  year      = {2026},
  publisher = {Association for Computing Machinery},
  address   = {New York, NY, USA},
  doi       = {10.1145/3772318.3791059},
  url       = {https://doi.org/10.1145/3772318.3791059}
}

@inproceedings{10.1145/3551349.3556903,
  author    = {Eghbali, Aryaz and Pradel, Michael},
  title     = {{CrystalBLEU}: Precisely and Efficiently Measuring the Similarity of Code},
  booktitle = {Proceedings of the 37th IEEE/ACM International Conference on Automated Software Engineering},
  series    = {ASE '22},
  articleno = {28},
  year      = {2023},
  publisher = {Association for Computing Machinery},
  address   = {New York, NY, USA},
  doi       = {10.1145/3551349.3556903},
  url       = {https://doi.org/10.1145/3551349.3556903}
}

@inproceedings{ostmeier-etal-2024-green,
  title     = {{GREEN}: Generative Radiology Report Evaluation and Error Notation},
  author    = {Ostmeier, Sophie and Xu, Justin and Chen, Zhihong and Varma, Maya and Blankemeier, Louis and Bluethgen, Christian and Michalson, Arne Edward and Moseley, Michael and Langlotz, Curtis and Chaudhari, Akshay S. and Delbrouck, Jean-Benoit},
  editor    = {Al-Onaizan, Yaser and Bansal, Mohit and Chen, Yun-Nung},
  booktitle = {Findings of the Association for Computational Linguistics: EMNLP 2024},
  month     = nov,
  year      = {2024},
  address   = {Miami, Florida, USA},
  publisher = {Association for Computational Linguistics},
  pages     = {374--390},
  doi       = {10.18653/v1/2024.findings-emnlp.21},
  url       = {https://aclanthology.org/2024.findings-emnlp.21/}
}

@article{chen2021codex,
  title   = {Evaluating Large Language Models Trained on Code},
  author  = {Chen, Mark and Tworek, Jerry and Jun, Heewoo and Yuan, Qiming and Pinto, Henrique Ponde de Oliveira and Kaplan, Jared and Edwards, Harri and Burda, Yuri and Joseph, Nicholas and Brockman, Greg and others},
  journal = {arXiv preprint arXiv:2107.03374},
  year    = {2021},
  url     = {https://arxiv.org/abs/2107.03374}
}

@inproceedings{wang2023grammar,
  author    = {Wang, Bailin and Wang, Zi and Wang, Xuezhi and Cao, Yuan and Saurous, Rif A. and Kim, Yoon},
  title     = {Grammar Prompting for Domain-Specific Language Generation with Large Language Models},
  booktitle = {Proceedings of the 37th International Conference on Neural Information Processing Systems},
  series    = {NIPS '23},
  articleno = {2837},
  year      = {2023},
  publisher = {Curran Associates Inc.},
  address   = {Red Hook, NY, USA}
}

@inproceedings{NEURIPS2025_APOLLO,
  author    = {Ospanov, Azim and Farnia, Farzan and Mohit, Roozbeh},
  title     = {{APOLLO}: Automated {LLM} and {Lean} Collaboration for Advanced Formal Reasoning},
  booktitle = {Advances in Neural Information Processing Systems},
  editor    = {Belgrave, D. and Zhang, C. and Lin, H. and Pascanu, R. and Koniusz, P. and Ghassemi, M. and Chen, N.},
  volume    = {38},
  pages     = {41599--41633},
  year      = {2025},
  publisher = {Curran Associates, Inc.},
  doi       = {10.52202/085713-1389},
  url       = {https://proceedings.neurips.cc/paper_files/paper/2025/file/3b77109ad4dd4ba82d07cacd4b24207e-Paper-Conference.pdf}
}

@misc{caron2021emergingpropertiesselfsupervisedvision,
  title         = {Emerging Properties in Self-Supervised Vision Transformers},
  author        = {Caron, Mathilde and Touvron, Hugo and Misra, Ishan and J{\'e}gou, Herv{\'e} and Mairal, Julien and Bojanowski, Piotr and Joulin, Armand},
  year          = {2021},
  eprint        = {2104.14294},
  archivePrefix = {arXiv},
  primaryClass  = {cs.CV},
  url           = {https://arxiv.org/abs/2104.14294}
}

@misc{oquab2024dinov2learningrobustvisual,
  title         = {{DINOv2}: Learning Robust Visual Features without Supervision},
  author        = {Oquab, Maxime and Darcet, Timoth{\'e}e and Moutakanni, Th{\'e}o and Vo, Huy and Szafraniec, Marc and Khalidov, Vasil and Fernandez, Pierre and Haziza, Daniel and Massa, Francisco and El-Nouby, Alaaeldin and Assran, Mahmoud and Ballas, Nicolas and Galuba, Wojciech and Howes, Russell and Huang, Po-Yao and Li, Shang-Wen and Misra, Ishan and Rabbat, Michael and Sharma, Vasu and Synnaeve, Gabriel and Xu, Hu and Jegou, Herv{\'e} and Mairal, Julien and Labatut, Patrick and Joulin, Armand and Bojanowski, Piotr},
  year          = {2024},
  eprint        = {2304.07193},
  archivePrefix = {arXiv},
  primaryClass  = {cs.CV},
  url           = {https://arxiv.org/abs/2304.07193}
}

@inproceedings{9879898,
  author    = {Tumanyan, Narek and Bar-Tal, Omer and Bagon, Shai and Dekel, Tali},
  title     = {Splicing {ViT} Features for Semantic Appearance Transfer},
  booktitle = {2022 IEEE/CVF Conference on Computer Vision and Pattern Recognition (CVPR)},
  year      = {2022},
  pages     = {10738--10747},
  publisher = {IEEE Computer Society},
  address   = {Los Alamitos, CA, USA},
  doi       = {10.1109/CVPR52688.2022.01048}
}

@misc{wightman_vit_small_dino,
  author       = {Wightman, Ross},
  title        = {{vit\_small\_patch16\_224.dino}},
  year         = {2022},
  month        = dec,
  publisher    = {Hugging Face},
  howpublished = {\url{https://huggingface.co/timm/vit_small_patch16_224.dino}},
  note         = {Model weights, distributed through timm. Initial commit December 22, 2022}
}

@misc{wightman_vit_small_dinov2,
  author       = {Wightman, Ross},
  title        = {{vit\_small\_patch14\_dinov2.lvd142m}},
  year         = {2023},
  month        = may,
  publisher    = {Hugging Face},
  howpublished = {\url{https://huggingface.co/timm/vit_small_patch14_dinov2.lvd142m}},
  note         = {Model weights, distributed through timm. Initial commit May 9, 2023}
}

@misc{rahmanzadehgervi2024vision,
  author        = {Rahmanzadehgervi, Pooyan and Bolton, Logan and Taesiri, Mohammad Reza and Nguyen, Anh Totti},
  title         = {Vision Language Models Are Blind: Failing to Translate Detailed Visual Features into Words},
  year          = {2024},
  month         = jul,
  eprint        = {2407.06581},
  archivePrefix = {arXiv},
  primaryClass  = {cs.AI},
  url           = {https://arxiv.org/abs/2407.06581},
  note          = {Version 6, revised March 27, 2025}
}

@inproceedings{vlmsarebiased,
  title     = {Vision Language Models Are Biased},
  author    = {Vo, An and Nguyen, Khai-Nguyen and Taesiri, Mohammad Reza and Dang, Vy Tuong and Nguyen, Anh Totti and Kim, Daeyoung},
  booktitle = {The Fourteenth International Conference on Learning Representations (ICLR)},
  year      = {2026},
  url       = {https://arxiv.org/abs/2505.23941}
}

@inproceedings{shahgir2026vlms,
  author    = {Shahgir, Haz Sameen and Chen, Xiaofu and Fu, Yu and Shayegani, Erfan and Abu-Ghazaleh, Nael and Kementchedjhieva, Yova and Dong, Yue},
  title     = {{VLMs} Need Words: Vision Language Models Ignore Visual Detail in Favor of Semantic Anchors},
  booktitle = {Proceedings of the Third Conference on Language Modeling (COLM 2026)},
  year      = {2026}
}

@inproceedings{wang2026compose,
  author    = {Wang, Yucheng and Hou, Yifan and Javadov, Aydin and Akhtar, Mubashara and Sachan, Mrinmaya},
  title     = {Compose and Fuse: Revisiting the Foundational Bottlenecks in Multimodal Reasoning},
  booktitle = {International Conference on Learning Representations (ICLR 2026)},
  year      = {2026}
}

@misc{fan2026tinted,
  author        = {Fan, Wan-Cyuan and Luo, Jiayun and Kutscher, Declan and Sigal, Leonid and Gupta, Ritwik},
  title         = {Tinted Frames: Question Framing Blinds Vision-Language Models},
  year          = {2026},
  month         = mar,
  eprint        = {2603.19203},
  archivePrefix = {arXiv},
  primaryClass  = {cs.CV},
  doi           = {10.48550/arXiv.2603.19203},
  note          = {Preprint, v2 revised March 20, 2026}
}

@inproceedings{wang-etal-2026-reasoning-benchmark,
  title     = {Your Reasoning Benchmark May Not Test Reasoning: Revealing Perception Bottleneck in Abstract Reasoning Benchmarks},
  author    = {Wang, Xinhe and Huang, Jin and Zhang, Xingjian and Wang, Tianhao and Ma, Jiaqi W.},
  editor    = {Liakata, Maria and Moreira, Viviane P. and Zhang, Jiajun and Jurgens, David},
  booktitle = {Proceedings of the 64th Annual Meeting of the {A}ssociation for {C}omputational {L}inguistics (Volume 1: Long Papers)},
  month     = jul,
  year      = {2026},
  address   = {San Diego, California, United States},
  publisher = {Association for Computational Linguistics},
  pages     = {18113--18124},
  doi       = {10.18653/v1/2026.acl-long.826},
  url       = {https://aclanthology.org/2026.acl-long.826/}
}

@inproceedings{wu2026seeing,
  author    = {Wu, Juncheng and Chen, Hardy and Tu, Haoqin and Tang, Xianfeng and Shi, Freda and Liu, Hui and Lu, Hanqing and Xie, Cihang and Zhou, Yuyin},
  title     = {From Seeing to Thinking: Decoupling Perception and Reasoning Improves Post-Training of Vision-Language Models},
  booktitle = {Proceedings of the 43rd International Conference on Machine Learning},
  year      = {2026}
}

@inproceedings{li2026unfolding,
  author    = {Li, Linjie and Bigverdi, Mahtab and Gu, Jiawei and Ma, Zixian and Yang, Yinuo and Li, Ziang and Choi, Yejin and Krishna, Ranjay},
  title     = {Unfolding Spatial Cognition: Evaluating Multimodal Models on Visual Simulations},
  booktitle = {International Conference on Learning Representations (ICLR 2026)},
  year      = {2026},
  note      = {Poster}
}

@inproceedings{zhang2026spinbench,
  author    = {Zhang, Yuyou and Corcodel, Radu and Hori, Chiori and Cherian, Anoop and Zhao, Ding},
  title     = {{SpinBench}: Perspective and Rotation as a Lens on Spatial Reasoning in {VLMs}},
  booktitle = {International Conference on Learning Representations (ICLR 2026)},
  year      = {2026}
}

@inproceedings{pmlr-v235-bachmann24a,
  title     = {The Pitfalls of Next-Token Prediction},
  author    = {Bachmann, Gregor and Nagarajan, Vaishnavh},
  booktitle = {Proceedings of the 41st International Conference on Machine Learning},
  series    = {Proceedings of Machine Learning Research},
  volume    = {235},
  pages     = {2296--2318},
  year      = {2024},
  editor    = {Salakhutdinov, Ruslan and Kolter, Zico and Heller, Katherine and Weller, Adrian and Oliver, Nuria and Scarlett, Jonathan and Berkenkamp, Felix},
  publisher = {PMLR},
  url       = {https://proceedings.mlr.press/v235/bachmann24a.html}
}

@inproceedings{ye2025beyond,
  author    = {Ye, Jiacheng and Gao, Jiahui and Gong, Shansan and Zheng, Lin and Jiang, Xin and Li, Zhenguo and Kong, Lingpeng},
  title     = {Beyond Autoregression: Discrete Diffusion for Complex Reasoning and Planning},
  booktitle = {The Thirteenth International Conference on Learning Representations (ICLR)},
  year      = {2025}
}

@article{mccoy2024embers,
  author  = {McCoy, R. Thomas and Yao, Shunyu and Friedman, Dan and Hardy, Mathew D. and Griffiths, Thomas L.},
  title   = {Embers of Autoregression Show How Large Language Models Are Shaped by the Problem They Are Trained to Solve},
  journal = {Proceedings of the National Academy of Sciences},
  volume  = {121},
  number  = {41},
  pages   = {e2322420121},
  year    = {2024},
  doi     = {10.1073/pnas.2322420121}
}

@misc{petty2026finegrainedhierarchybenchmarkstate,
  title         = {A Fine-Grained Hierarchy and Benchmark for State Tracking in Language Models},
  author        = {Petty, Jackson and Merrill, William and Sabharwal, Ashish},
  year          = {2026},
  eprint        = {2611.03094},
  archivePrefix = {arXiv},
  primaryClass  = {cs.AI},
  url           = {https://arxiv.org/abs/2611.03094}
}

@inproceedings{snowballing,
  title     = {How Language Model Hallucinations Can Snowball},
  author    = {Zhang, Muru and Press, Ofir and Merrill, William and Liu, Alisa and Smith, Noah A.},
  booktitle = {Proceedings of the 41st International Conference on Machine Learning},
  series    = {Proceedings of Machine Learning Research},
  volume    = {235},
  pages     = {59670--59684},
  year      = {2024},
  editor    = {Salakhutdinov, Ruslan and Kolter, Zico and Heller, Katherine and Weller, Adrian and Oliver, Nuria and Scarlett, Jonathan and Berkenkamp, Felix},
  publisher = {PMLR},
  url       = {https://proceedings.mlr.press/v235/zhang24ay.html}
}

@inproceedings{handa2025actionreasoningbench,
  author    = {Handa, Divij and Dolin, Pavel and Kumbhar, Shrinidhi and Son, Tran and Baral, Chitta},
  title     = {{ActionReasoningBench}: Reasoning about Actions with and without Ramification Constraints},
  booktitle = {International Conference on Learning Representations (ICLR 2025)},
  year      = {2025},
  note      = {Poster}
}

@inproceedings{mancoridis2025potemkin,
  author    = {Mancoridis, Marina and Weeks, Bec and Vafa, Keyon and Mullainathan, Sendhil},
  title     = {{Potemkin} Understanding in Large Language Models},
  booktitle = {Proceedings of the 42nd International Conference on Machine Learning (ICML 2025)},
  year      = {2025},
  note      = {Poster}
}

@inproceedings{mirzadeh2025gsmsymbolic,
  author    = {Mirzadeh, Iman and Alizadeh, Keivan and Shahrokhi, Hooman and Tuzel, Oncel and Bengio, Samy and Farajtabar, Mehrdad},
  title     = {{GSM-Symbolic}: Understanding the Limitations of Mathematical Reasoning in Large Language Models},
  booktitle = {International Conference on Learning Representations (ICLR 2025)},
  year      = {2025}
}

@misc{gdm2026gemini37flash,
  author       = {{Google DeepMind}},
  title        = {{Gemini 3.7 Flash} Model Card},
  year         = {2026},
  month        = aug,
  howpublished = {\url{https://deepmind.google/models/model-cards/gemini-3-7-flash/}},
  note         = {Published 13 August 2026}
}

@misc{openai2026gpt56,
  author       = {{OpenAI}},
  title        = {{GPT-5.6}: Frontier Intelligence That Scales with Your Ambition},
  year         = {2026},
  month        = jul,
  howpublished = {\url{https://openai.com/index/gpt-5-6/}},
  note         = {OpenAI blog post, July 9, 2026}
}

@misc{moonshot2026kimik3,
  author       = {{Moonshot AI}},
  title        = {{Kimi-K3}},
  year         = {2026},
  month        = jul,
  publisher    = {Hugging Face},
  howpublished = {\url{https://huggingface.co/moonshotai/Kimi-K3}},
  note         = {Model weights. Initial commit July 27, 2026}
}

@misc{thinkingmachines2026inkling,
  author       = {{Thinking Machines Lab}},
  title        = {{Inkling} Model Card},
  year         = {2026},
  month        = jul,
  howpublished = {\url{https://thinkingmachines.ai/model-card/inkling/}},
  note         = {Released July 15, 2026. Apache 2.0 license}
}

@misc{thinkingmachines2026tinker,
  author       = {{Thinking Machines Lab}},
  title        = {{Tinker}},
  year         = {2026},
  howpublished = {\url{https://thinkingmachines.ai/tinker/}},
  note         = {Product page. Accessed October 28, 2026}
}

@techreport{anthropic2026sonnet5,
  author      = {{Anthropic}},
  title       = {System Card: {Claude Sonnet 5}},
  institution = {Anthropic},
  type        = {System card},
  year        = {2026},
  month       = jun,
  url         = {https://assets.claude.com/283ef97c476cf442c91d9a37d5b214242a55bb92.pdf},
  note        = {Changelog updated July 10, 2026}
}

@misc{qwen2026qwen35model,
  author       = {{Qwen}},
  title        = {{Qwen3.5-4B}},
  year         = {2026},
  month        = feb,
  publisher    = {Hugging Face},
  howpublished = {\url{https://huggingface.co/Qwen/Qwen3.5-4B}},
  note         = {Model weights. Initial commit February 27, 2026}
}

@misc{qwen2026qwen359b,
  author       = {{Qwen}},
  title        = {{Qwen3.5-9B}},
  year         = {2026},
  month        = feb,
  publisher    = {Hugging Face},
  howpublished = {\url{https://huggingface.co/Qwen/Qwen3.5-9B}},
  note         = {Model weights. Initial commit February 27, 2026}
}

@misc{qwen2026qwen36,
  author       = {{Qwen}},
  title        = {{Qwen3.6-35B-A3B}},
  year         = {2026},
  month        = apr,
  publisher    = {Hugging Face},
  howpublished = {\url{https://huggingface.co/Qwen/Qwen3.6-35B-A3B}},
  note         = {Model weights. Initial commit April 15, 2026}
}

@misc{qwen2024qwen2vl7b,
  author       = {{Qwen}},
  title        = {{Qwen2-VL-7B-Instruct}},
  year         = {2024},
  month        = aug,
  publisher    = {Hugging Face},
  howpublished = {\url{https://huggingface.co/Qwen/Qwen2-VL-7B-Instruct}},
  note         = {Model weights. Initial commit August 2024}
}

@misc{meta2024llama33,
  author       = {{Meta AI}},
  title        = {{Llama-3.3-70B-Instruct}},
  year         = {2024},
  month        = dec,
  publisher    = {Hugging Face},
  howpublished = {\url{https://huggingface.co/meta-llama/Llama-3.3-70B-Instruct}},
  note         = {Model weights and model card. Released December 6, 2024}
}

\appendix

\section{Overview of the Appendix}
\label{app:app-overview}
The appendix is organised as follows:
\begin{itemize}[itemsep=0ex, topsep=0ex, parsep=0pt]
    \item \textit{Section B} offers a simple formalisation of crochet, which we refer to throughout this paper.
    \item \textit{Section C} provides details of our experiments with the lexical overlap metrics underpinning CrochetBench Task C (BLEU, ROUGE and ChrF) as discussed in \S\ref{sec:benchmarks}. 
    \item \textit{Section D} explains our experiments with the metrics used in CrochetBench Task D, namely CrochetPARADE DSL Valid Pattern Rate and DINO [CLS] render-photo similarity. We refer to these in \S\ref{sec:benchmarks}.
    \item \textit{Section E} covers the FoundationChain targets, execution rules, sampling settings and scaffold referred to in \S\ref{sec:foundationchain}.
    \item \textit{Section F} contains sample outputs for each model across all of the generation stages as described in \S\ref{sec:results}.
    \item \textit{Section G} includes additional details of the results \S\ref{sec:results}.
\end{itemize}

\section{Crochet: A Simple Formalisation}
\label{app:graph}

\paragraph{What are patterns for?} People (crocheters) follow a pattern to make an object with a given shape: a beanie hat that fits one's head, a granny square that can be joined together with others into a blanket, an amigurumi toy that looks like a teddy bear, a section of the hyperbolic plane \cite{henderson2001crocheting}, and so on. The pattern does not describe the shape directly. Instead, it lays out an ordered set of actions crocheters follow, and the shape emerges from carrying them out.

\paragraph{Patterns as sequences of actions.} A pattern specifies a
sequence of actions $A_1, \ldots, A_n$ (individual stitches, and operations such as chaining, turning, joining and fastening off), grouped into rows or rounds. Patterns express this sequence compactly through repeats, and rely on conventions and tacit knowledge that the crocheter supplies, including the terminology itself (a UK double crochet is a US single crochet; we use US terms throughout).

\paragraph{Patterns as graphs of stitches.} Carrying out the sequence builds a graph $G = (V, E)$ whose nodes are stitches (\citealp[following][]{tassev2026layout, representingstitch, 10.2312:exw.20251057, amigo}). Each stitch is joined to the stitch made just before it and to whatever part of the existing piece it is worked into. This is often the top of a stitch in the previous row or round, but a crocheter can insert the hook anywhere in the fabric: into any earlier stitch, around its post, through a single loop (e.g., front or back loop only, which produces ridges and creases), or into a space between stitches. To increase is to work several stitches into the same place. To decrease is to work one stitch across several, consuming multiple positions at once. Edges between consecutive stitches run along the strand, with lengths roughly equal to a stitch’s width; edges connecting the stitch to the position it is worked into have lengths roughly equal to its height. Both depend on the stitch type, yarn, hook and tension (e.g., a double crochet stitch is roughly twice the height of a single crochet stitch, but similar in width), and edges also record where and how each stitch was attached. The ratio between the two governs curvature: a flat circle, for instance, must add roughly $2\pi H/W$ stitches per round, where $H$ and $W$ are stitch height and width \citep{kekkonen2024bour}. 

\paragraph{States and transitions.} Starting from $S_0$, a slip knot or starting loop on the hook, each action $A_t$ takes the piece from state $S_{t-1}$ to $S_t = f(S_{t-1}, A_t)$, where a state $S_t = (G_t, x_t, h_t)$ comprises the stitches made so far, $G_t \subseteq G$; the shape they have taken, an assignment $x_t$ of 3D positions to the nodes of $G_t$ \citep[cf.][]{agarwal2026aiunderstandlanguageorigami}; and the working position $h_t$ (the live loop, the next stitch in the current row or round, and which side of the piece faces the crocheter). The finished object is the final state $S_n$, with $G_n = G$.

\paragraph{Shape.} The shape the piece takes at each step is a low-energy configuration of some energy $\mathcal{U}$, which depends on the stitches, where and how they were worked, the yarn, the crocheter's own tension (how tightly they hold the yarn and work it into the other stitches) and how the piece itself is handled. Many shapes are consistent with the same graph, and since the physical $\mathcal{U}$ is unknown, simulators must approximate it \citep[e.g.][]{tassev2026layout, 10.2312:exw.20251057, amigo}. Moreover, many graphs yield essentially the same shape. This is why in work with computational crochet, the inverse direction (objects into patterns) has so far proved more fruitful \citep[e.g.,][]{amigo, representingstitch, 10.2312:pg.20261024} than the forward one (patterns into objects) which remains largely an open problem \citep[e.g.,][]{10.2312:exw.20251057, 10.1145/3563835.3567657, tassev2026layout, storck2022topology}.

\paragraph{Sequentiality and path dependence.} Crochet is strictly sequential: each action $A_t$ is applied to the state $S_{t-1}$ produced by all previous actions, so an error at step $t$ propagates through every subsequent state. It can also only be undone in reverse: within a continuous strand, the fabric is worked from one live loop, so undoing a stitch means undoing every stitch made since. Moreover, the shape is path-dependent: each new shape $x_t$ is reached by relaxing from $x_{t-1}$ with the new stitch added, so the piece may settle into a local rather than a global minimum of $\mathcal{U}$, and the route taken can affect where it ends up. Tension also varies within one sitting and between crocheters, so $f$ is best read as an idealisation of a process that is, in practice, very noisy.

\paragraph{The task as inversion.} Our testbed, {FoundationChain}, asks models to run this process backwards. Given an image of a finished object, a model must write a text which prescribes a series of actions $A_1, \ldots, A_n$ that build a graph $G$ that settles into a shape $x_n$ matching the target. The target's arrangement of stitches, their types, heights and texture largely determine $G$ and the order in which it was built, but not how that sequence is expressed: the same pattern can be written in many different ways. A correct pattern therefore may not strongly resemble a particular reference text, but it must produce the right object when executed.

\paragraph{Errors by layer.} Our error taxonomy (\S\ref{sec:foundationchain}) locates failures within this structure. \emph{Hallucination} arises in interpretation: the VLM attributes features to $x_n$ that it does not have. \emph{Count}, \emph{state tracking} and \emph{ambiguity} errors concern actions and state: stated counts misdescribe $G_t$, $A_t$ is incompatible with $G_{t-1}$ or $h_t$, or an instruction does not determine a unique $A_t$ given the current state. \emph{Stitch} errors concern the nodes of $G$: the wrong stitches and thus edge lengths. \emph{Construction} errors pertain to the order in which $G$ is built and embedded: crochet objects can be made horizontally \textit{in rows}, turning the work over at the end of each row, which typically produces flat pieces that grow `upwards’. They can also be worked \textit{in the round}, whereby objects grow outward from a central ring without turning, as for circles, polygons and most 3D shapes. Each method can be used to create some geometries but not others, so choosing incorrectly can make the target shape impossible to achieve. \emph{Geometry} and \emph{size} errors concern the shape: the finished $x_n$ differs from the target in form or extent.

\section{CrochetBench Task C: BLEU, ChrF and ROUGE-L}
\label{app:ngrams}

We score each of the 6,084 reference patterns in the CrochetBench Task C corpus against the reference of a different, randomly assigned pattern (no pattern is paired with itself), using CrochetBench's own implementations of BLEU, ROUGE-L and ChrF (commit \pat{4f834d5}, 9 September 2026; \citealp{li2026crochetbench_code}). Repeating this over five seeds gives BLEU $4.25 \pm 0.06$, ROUGE-L $20.58 \pm 0.06$ and ChrF $26.43 \pm 0.07$ (mean $\pm$ s.d.\ over pairings). Pairs themselves are more variable (median/p90: BLEU 2.64/10.62, ROUGE-L 20.48/29.03, ChrF 27.12/38.33).

Table~\ref{tab:shared-ngrams} lists a selection of the word $n$-grams most often shared between unrelated patterns in one such pairing, tokenised on whitespace as is the case in CrochetBench's implementation of BLEU. Of the $n$-grams in each pattern, 27.1\% of unigrams, 9.9\% of bigrams, 4.4\% of trigrams and 2.2\% of 4-grams can also be found in its unrelated partner and 82\% of unrelated pairs share at least one 4-gram. The same can be seen with ChrF, which instead compares character $n$-grams ($n = 1$ to $6$) with whitespace removed. The longest shared character sequences are many of the same generic phrases. \texttt{Fasten} occurs in 94\% of unrelated pairs, \texttt{innext} in 80\% and \texttt{ineach} in 77\%. ROUGE-L instead rewards the longest common subsequence of tokens (LCS). Across a sample of 1,000 unrelated pairs, we find a median LCS of 123 tokens versus a median total pattern length of 552 tokens (recall 25.6\%, precision 26.1\%). Fifteen tokens (\texttt{in}, \texttt{ch}, \texttt{sc}, \texttt{1}, \texttt{next}, \texttt{2}, \texttt{of}, \texttt{st}, \texttt{dc}, \texttt{row}, \texttt{each},
\texttt{to}, \texttt{3},
\texttt{with},
\texttt{from}) account for 76.2\% of this overlap. In each case, these metrics effectively just count stock phrases and numerals common to all crochet patterns.

\begin{table}[h]
\centering
\small
\begin{tabular}{llr}
\hline
$n$ & Shared $n$-gram & Pairs (\%) \\
\hline
1 & \texttt{in}                        & 99.9 \\
1 & \texttt{of}                        & 97.0 \\
1 & \texttt{ch}                        & 95.5 \\
1 & \texttt{2}                         & 90.9 \\
\hline
2 & \texttt{in next}                   & 79.8 \\
2 & \texttt{in each}                   & 77.3 \\
2 & \texttt{Fasten off.}               & 76.7 \\
2 & \texttt{from *}                    & 60.1 \\
\hline
3 & \texttt{sc in each}                & 40.0 \\
3 & \texttt{in 2nd ch}                 & 26.6 \\
3 & \texttt{in each of}                & 23.6 \\
3 & \texttt{dc in next}                & 21.0 \\
3 & \texttt{Rep from *}                & 20.9 \\
3 & \texttt{with sl st}                & 20.3 \\
\hline
4 & \texttt{in 2nd ch from}            & 25.4 \\
4 & \texttt{in each of next}           & 20.6 \\
4 & \texttt{with sl st to}             & 16.7 \\
4 & \texttt{sc in each sc}             & 13.0 \\
4 & \texttt{Join with sl st}           & 11.6 \\
4 & \texttt{each of next 2}            & 10.4 \\
\hline
5 & \texttt{sc in 2nd ch from}         & 13.0 \\
5 & \texttt{Join with sl st to}        & 10.3 \\
5 & \texttt{Ch 1. 1 sc in}             & 10.1 \\
5 & \texttt{in each of next 2}         &  9.5 \\
5 & \texttt{sc in each of next}        &  6.7 \\
5 & \texttt{to end of row. Turn.}      &  6.0 \\
\hline
6 & \texttt{Join with sl st to first}  &  8.2 \\
6 & \texttt{sc in 2nd ch from hook.}   &  6.4 \\
6 & \texttt{Ch 1. 1 sc in each}        &  5.3 \\
6 & \texttt{1 sc in each of next}      &  4.3 \\
6 & \texttt{in each ch to end of}      &  3.7 \\
6 & \texttt{row: Ch 1. 1 sc in}        &  3.5 \\
\hline
\end{tabular}
\caption{Word $n$-grams shared between a pattern and an unrelated pattern (one random pairing of 6,084 patterns;
tokenised on whitespace and case-sensitive, as in CrochetBench's BLEU). Rows are selected from the ten most
shared $n$-grams of each length (four for $n \le 2$, six for $n \ge 3$); a strict top-$k$ is dominated by
overlapping fragments of the same few phrases (e.g.\ \texttt{in 2nd ch}, \texttt{2nd ch from}, \texttt{sc in 2nd ch}),
so we show a range of phrases instead. ``Pairs'' is the percentage of pairs in which the $n$-gram occurs in both patterns.}
\label{tab:shared-ngrams}
\end{table}

For each of the 6,084 reference patterns, we also replace every integer (every maximal run of digits, apart from the digits inside increase or decrease names such as \texttt{sc2tog}) with a different integer drawn uniformly from $1$ to $\max(10, 2v)$, excluding the original value $v$. This changes all 6,084 patterns (136.7 integers per pattern on average), and perturbs every row number, stitch count, increase, decrease and numbered repeat. Scoring each perturbed copy against its unaltered original gives BLEU $64.23 \pm 0.01$, ROUGE-L $82.47 \pm 0.00$ and ChrF $85.09 \pm 0.01$ (mean $\pm$ s.d.\ over five seeds), far above every model CrochetBench evaluates. Because stitch names and connective phrases surrounding numerals still match, the metrics barely register that the pattern now describes an entirely different object.

\section{CrochetBench Task D: DSL Valid Pattern Rate and DINO Similarity}
\label{app:taskd}

CrochetBench Task D \citep{crochetbench} tests whether models can translate a human-authored natural-language (NL) crochet pattern into the CrochetPARADE DSL \citep{tassev2023crochetparade}. It has two variants. In $D_{proj}$ (100 examples), the model is given a complete NL pattern from Yarnspirations \cite{spinrite2026yarnspirations} together with a product image and must generate an entire CrochetPARADE program for it. The generated program is scored first on whether it compiles (Valid Pattern Rate; VPR) and then on the DINO similarity between a 3D wireframe render of it and a photo of the target object (\S\ref{app:dino}). In $D_{step}$, the model is given the first few instructions of a pattern alongside their correct DSL translations, and must translate the next NL instruction into DSL. $D_{step}$ is scored solely on whether the DSL program compiles successfully. 

Our analysis of Task D focuses on $D_{proj}$, as it is the only variant that is compared to a ground-truth object. We use CrochetBench’s code and data at commit \pat{4f834d5} (9 September 2026; \citealp{li2026crochetbench_code}). One $D_{proj}$ record (\pat{RHC0542-016837M: Home Spa Bath Mat}) lacks NL instructions, so the maximum attainable VPR is 99\%. We generated translations using CrochetBench’s code and data from commit \pat{930b196} (12 July 2026), and ran the DINO similarity evaluation using commit \pat{4f834d5} (9 September 2026), the first to include code for this metric. The Task D data is identical in both.

Unlike CrochetBench, we provide models only with the natural-language pattern and omit the product image. CrochetBench suggests the image may help models translate aspects of patterns that are under-specified in text, but this does not affect our conclusions. This is because VPR only measures whether DSL programs are \emph{syntactically valid}, not whether they are \emph{correct}. Moreover we significantly exceed the VPR scores CrochetBench reports with minimal changes to the harness. Our analysis of DINO similarity is also unaffected, as we find renders of unrelated items are more similar to one another than matched render-photo pairs are. Moreover, our matched-pair scores are themselves consistent with those CrochetBench reports (\S\ref{app:dino}).

\subsection{Models and Sampling Settings}
\label{app:crochetbench-sampling}

We evaluate eight models on Task D-proj: Gemini 3.7 Flash \cite{gdm2026gemini37flash}, GPT-5.6 Terra \cite{openai2026gpt56}, Kimi K3 \cite{moonshot2026kimik3},  Inkling \cite{thinkingmachines2026inkling}, Claude Sonnet 5  \cite{anthropic2026sonnet5}, Qwen3.6-35B-A3B \cite{qwen2026qwen36}, Qwen3.5-9B \cite{qwen2026qwen359b}and Qwen3.5-4B \cite{qwen2026qwen35model}, as well as our fine-tuned Qwen3.5-4B (\S\ref{app:finetune}). Gemini 3.7 Flash, GPT-5.6 Terra and Claude Sonnet 5  were accessed through their providers’ APIs, all other models were accessed through the Doubleword Batch API. We generated all outputs between the 5th of August and 1st of September 2026. We use temperature 0 with extended reasoning disabled where possible (Gemini 3.7 Flash can only be set to `low’), and generate one translation per pattern. Maximum output length is 2048 tokens.

\subsection{Results: Valid Pattern Rate}
\label{app:crochetbench-sampling}
CrochetBench reports just 5--8\% VPR for the strongest models it tests, and takes this as evidence of severe failures in procedural reasoning \citep[p.~13235]{crochetbench}. These scores are in line with our own: under CrochetBench’s prompt (\ref{app:crochetbench-prompt}), no model we test exceeds 7\% VPR.

However, we find VPR depends heavily on the harness used and can be raised substantially with minimal effort (Table~\ref{tab:vpr}). Providing a summary of CrochetPARADE’s syntax (\ref{app:enhanced-prompt}) raises VPR as high as 28\%, while adding one round of compiler feedback and two simple post-processors (\ref{app:compilerfeedback-prompt}) raises it to as much as 64\%. The most frequent errors CrochetBench reports, undefined stitches and unbalanced brackets, are exactly those these changes address. Fine-tuning has even greater effect, taking the worst performing model, Qwen3.5-4B from 1\% to 73\% VPR under CrochetBench's prompt (\S\ref{app:finetune}). As such, we argue VPR confounds unfamiliarity with the CrochetPARADE DSL with poor crochet ability.

\begin{table}[t]
\centering
\small
\setlength{\tabcolsep}{4pt}
\begin{tabular}{lccc}
\hline
      & CB     & Enhanced & + post-proc. \\
Model & prompt & prompt   & + 1 retry \\
\hline
Gemini 3.7 Flash       &  7 & 28 & 64 \\
GPT-5.6 Terra           &  0 & 21 & 40 \\
Kimi K3                  &  1 & 20 & 34 \\
Inkling                 &  3 & 22 & 29 \\
Claude Sonnet 5        &  1 & 12 & 20 \\
Qwen3.6-35B-A3B         &  0 &  7 & 10 \\
Qwen3.5-9B              &  0 &  4 &  7 \\
Qwen3.5-4B (base)       &  1 &  2 &  2 \\
\hline
Qwen3.5-4B (fine-tuned) & 73 & 75 & 77 \\
\hline
\end{tabular}
\caption{Valid Pattern Rate (VPR) (\%) on 100 CrochetBench Task D-proj test patterns. The 'enhanced' prompt adds basic syntax guidance. The final column adds post-processing and one round of compiler feedback.}
\label{tab:vpr}
\end{table}

\subsection{CrochetBench Task D Prompt}
\label{app:crochetbench-prompt}
We use CrochetBench’s original Task D prompt, reproduced below. It gives minimal instructions but includes one long worked example: a complete pattern alongside its DSL translation.

\begin{promptbox}[title={System}]
You are a professional crochet pattern writer.
Convert instructions into compilable CrochetPARADE DSL code.
Output only the DSL code. No explanations, commentary, or extra text.
\end{promptbox}

\begin{promptbox}[title={User}]
Convert the following instructions into DSL:
Note:  Join with sl st to first sc at end of each rnd.
        Ch 2.
        **Rnd 1:** 6 sc in 2nd ch from hook. Join. (6 sc)
        **Rnd 2:** Ch 1. 2 sc in each sc around. Join. (12 sc)
        **Rnd 3:** Ch 1. _(2 sc in next sc, 1 sc in next sc)_ repeat around. End with 1 sc. Join. (18 sc)
        **Rnd 4:** Ch 1. _(2 sc in next sc, 1 sc in each of next 2 sc)_ repeat. End with 1 sc in last 2 sc. Join. (24 sc)
        **Rnd 5:** Ch 1. Sc in each sc around. Join. (24 sc)
        **Rnd 6:** Ch 1. _(2 sc in next sc, 1 sc in each of next 3 sc)_ repeat. End with 1 sc in last 3 sc. Join. (30 sc)
        **Rnds 7--8:** Repeat Rnd 5 (sc in each sc). Join. (30 sc each round)
        **Rnd 9:** Ch 1. **Working in back loops only**: _(2 sc in next sc, 1 sc in each of next 2 sc)_ repeat. End with 1 sc in last 2 sc. Join. (40 sc)
        **Rnd 10:** Ch 1. Sc in each sc around (both loops). Join. (40 sc)
        **Rnd 11:** Ch 1. _(2 sc in next sc, 1 sc in each of next 3 sc)_ repeat. End with 1 sc in last 3 sc. Join. (50 sc)
        **Finish:** Fasten off.
Expected DSL:

        ch.B
        sc@B.A,5sc@B,ss@A
        ch.A,sk,6sc2inc,ss@A
        ch.A,sk,[sc2inc,sc]*6,ss@A
        ch.A,sk,[sc2inc,2sc]*6,ss@A
        ch.A,sk,24sc,ss@A
        ch.A,sk,[sc2inc,3sc]*6,ss@A
        [ch.A,sk,30sc,ss@A
        ]*2
        ch.A,sk,[scbl,scbl@[@],2scbl]*10,ss@A
        ch.A,sk,40sc,ss@A
        ch.A,sk,[sc2inc,3sc]*10,ss@A

Convert the following instructions into DSL:
    {target NL instructions}
\end{promptbox}

\subsection{Our `Enhanced' NL-DSL Prompt}
\label{app:enhanced-prompt}
Our `enhanced' system prompt instead gives a longer summary of the DSL’s basic syntax and includes only one very short worked example.

\begin{promptbox}[title={System Prompt}]
Crochet compiler: convert NL crochet instructions to CrochetPARADE domain specific language (DSL). Output ONLY DSL, one row/round per line --- no prose, markdown, or stitch counts.

Token: <count><stitch>, count first --- 6sc, 10ch, 2sk
Stitches include: ring, ch, ss, sc, hdc, dc, tr, dtr, trtr
Use DEF: to define new stitches before first use.

Slip stitch = ss (never sl/slst).
Syntax:
- Skip stitches with sk.
- Increase and decrease with stitch_nameNinc, stitch_nameNtog
- Turn at end of row with turn
- # for comments
- Comma-separate stitches per line: 6ch, turn
- Repeats are with [group]*N or N*[group], N integer, brackets always balanced: [10sc,turn]*3
- Label a stitch with period and label letter e.g., .A when made to work into it later with @A (each @A needs an earlier .A)
- ring.R alone on its own line before any @R

Example --- NL: Ch 6. Row 1: sc in each ch across, turn. (6 sc)
DSL:
6ch, turn
6sc, turn
\end{promptbox}

\begin{promptbox}[title={User Prompt}]
Now translate this pattern into CrochetPARADE DSL:
NL:
{target NL pattern}
DSL:
\end{promptbox}

\subsection{Compiler Feedback and DSL Post-Processors}
\label{app:compilerfeedback-prompt}

DSL programs that fail to compile under the enhanced prompt are returned to the model once, together with the original pattern and the compiler error, using the following prompts.

\begin{promptbox}[title={System Prompt}]
You are a crochet compiler. You will be given a natural language crochet pattern, a previous DSL translation attempt and the compiler error it produced. Output only the corrected CrochetPARADE DSL with the error fixed.
\end{promptbox}

\begin{promptbox}[title={User Prompt}]
NL:
{target nl instructions}

Previous DSL attempt:
{previous non-compiling attempt}

Compiler error:
{compiler error text}

Output the corrected DSL only:
\end{promptbox}

For DSL programs that do not compile after the second attempt, we automatically apply two simple post-processors which together, increase VPR by between 0--6\%. They are: 
\begin{enumerate}[itemsep=0.5ex, topsep=0.5ex, parsep=0pt]
    \item Moving integer counts \emph{before} the stitch, as the DSL requires. This is one of the most common errors models make, probably because patterns typically put the count after the stitch (e.g., `chain 5' or `ch 5', `skip 2'). We rewrite \texttt{ch5}, \texttt{CH5} and \texttt{ch 5} as \texttt{5ch}, and \texttt{sk2} as \texttt{2sk}.
    \item As CrochetPARADE only allows turning to occur at the end of a row, if a bracketed repeat contains \texttt{turn} (e.g. \texttt{[12sc, turn]*4}), the bracket contents are expanded.
\end{enumerate}

Any programs that do not compile after two attempts and the post-processors are applied are counted as failures.

\subsection{Fine-tuning Qwen3.5-4B}
\label{app:finetune}

We fine-tune Qwen3.5-4B \cite{qwen2026qwen35model} with Thinking Machines' Tinker \cite{thinkingmachines2026tinker} supervised trainer to translate English-language crochet patterns into the CrochetPARADE DSL.

All patterns come from the CrochetBench Task~C training split used to finetune Qwen2-VL-7B \cite{qwen2024qwen2vl7b} in that paper \cite{li2026qwen2vl_crochetbench, crochetbench}, which contains 4{,}859 Yarnspirations \cite{spinrite2026yarnspirations} patterns. We carefully ensure that these do not include the 100 Task D test patterns. The training set contains 2{,}026 examples (2.72M tokens, 0.72M are target tokens), of which 1{,}949 were generated using Llama-3.3-70B-Versatile \cite{meta2024llama33}. 1{,}458 examples are straightforward English-to-DSL translations and the other 568 are auxiliary tasks built from failed translations. In 265 correction examples, the model receives a non-compiling program and its compiler error and must return a fixed program. In 131 detection examples, it must list the DSL rows whose stitch counts disagree with the NL pattern. In 95 trajectory examples, it must repair a program that compiles but has wrong per-row stitch counts. The remaining 77 examples use the correction input, but the target DSL comes from CrochetPARADE's own experimental deterministic NL-DSL translator \cite{tassev2023crochetparade}. 

We use LoRA rank 8, a learning rate of $1.5\times10^{-4}$ with a cosine schedule, batch size 16, 2 epochs and a maximum sequence length of 8{,}192 tokens. The longest example is 7{,}385 tokens, so no examples are truncated. Training is text-only with thinking disabled. The data build uses a fixed seed and checkpoints are saved every 25 steps.

All evaluations use greedy decoding. We select the step-175 checkpoint for its valid pattern rate (VPR) on the 100 CrochetBench Task~D test patterns, where it reaches 73\% under the CrochetBench prompt. Since our aim is to test how far familiarity with the DSL alone can raise VPR, we select on this set directly.

The model exhibits catastrophic forgetting, as it can only write CrochetPARADE DSL, even when asked for natural-language crochet patterns. When prompted to generate original patterns, including from the FoundationChain target images, the fine-tuned model produces severely degenerate outputs dominated by the same stitches repeated over and over, as shown below. For this reason, we do not use it in the subsequent Task D DINO similarity test, or to generate patterns for human execution in the rest of this paper. 

\begin{promptbox}[title={Fine-tuned Qwen3.5-4B: Pattern Generated for Granny Square Target}]
10ch, turn
(5sc, 5sc)*5, turn
(5sc, 5sc)*5, turn
(5sc, 5sc)*5, turn
(5sc, 5sc)*5, turn
(5sc, 5sc)*5, turn
(5sc, 5sc)*5, turn
(5sc, 5sc)*5, turn 
(5sc, 5sc)*5, turn 
(5sc, 5sc)*5, turn 
...
\end{promptbox}

\subsection{DINO Render-Photo Similarity}
\label{app:dino}

CrochetBench tests whether NL-CrochetPARADE DSL translations generated by models in Task D produce an output that resembles the target object by comparing a wireframe generated by CrochetPARADE's renderer to the item's product photo. 

We use the evaluation code provided by the CrochetBench authors, which loads the original DINO checkpoint \texttt{vit\_small\_patch16\_224.dino} \citep{caron2021emergingpropertiesselfsupervisedvision, wightman_vit_small_dino} and scores [CLS] token cosine similarity (Table~\ref{tab:dino-pooled}, left column). As the paper's citation for DINO is DINOv2 \citep{oquab2024dinov2learningrobustvisual}, we also repeat every analysis with \texttt{vit\_small\_patch14\_dinov2.lvd142m} \cite{wightman_vit_small_dinov2}, while keeping all other settings fixed (Table~\ref{tab:dino-pooled}, right column).

Only outputs that compile are rendered. The pool covers seven of the models used in our reproduction of Task D (\S\ref{app:taskd}): Gemini 3.7 Flash, GPT-5.6 Terra, Inkling, Kimi K3, Qwen3.6-35B-A3B, Qwen3.5-9B and Claude Sonnet 5. We exclude our fine-tuned Qwen3.5-4B. The base Qwen3.5-4B did not produce any renderable outputs. For each model and pattern we take the enhanced-prompt output that compiles after post-processing, from the first attempt or else the retry (the final column of Table~\ref{tab:vpr}). Of 204 compiling outputs, 134 (65.7\%) yielded a render before the timeout imposed by CrochetBench's implementation. Every render is scored against the product photo of every target with at least one render, giving 56 unique photos out of the 100 Task D patterns. Matched pairs are compared with mismatched pairs using $d = (\bar{s}_{\text{matched}} - \bar{s}_{\text{mismatched}}) /
\mathrm{sd}(s_{\text{mismatched}})$ and a 10{,}000-permutation $p$-value.

We hold the render–photo similarity matrix fixed and randomly permute which photo is treated as each render’s target (10,000 permutations), using the 56 items with at least one compiling, rendered output from the models we test. The matched–mismatched difference is statistically detectable given the number of pairs ($p = 0.007$), but negligible in size ($d = 0.11$): the correct photo’s mean rank is 26.4 of 56 (chance 28.5), and it is ranked first for 1.5\% of renders (chance 1.8\%). CrochetBench treats a [CLS] token cosine similarity of 0.6 or above as a good match. Our matched-pair scores are consistent with those it reports (0.143--0.215), and are far below this threshold.

Results with DINOv2 are mostly the same: the difference is again detectable but negligible ($p = 0.001$, $d = 0.12$), with a mean rank of 26.4 and a rank-1 rate of 2.2\%, while overall similarity is lower across the board (matched pairs score just 0.106 on average).

\begin{table}[t]
\centering
\small
\setlength{\tabcolsep}{4pt}
\begin{tabular}{lcc}
\hline
 & DINO & DINOv2 \\
\hline
Renders / photos & \multicolumn{2}{c}{134 / 56} \\
Matched pairs & 0.188 & 0.106 \\
Mismatched pairs & 0.176 & 0.095 \\
$d$ & 0.11 & 0.12 \\
Perm.\ $p$ & 0.007 & 0.001 \\
Rank-1 (chance 1.8\%) & 1.5\% & 2.2\% \\
Own rank (chance 28.5) & 26.42 & 26.36 \\
Render--render & 0.58 & 0.57 \\
Render--photo & 0.18 & 0.10 \\
Photo--photo & 0.27 & 0.20 \\
\hline
\end{tabular}
\caption{DINO and DINOv2 similarity (mean cosine similarity unless noted), renders vs.\ product photos. Ranks are out of 56 photos.}
\label{tab:dino-pooled}
\end{table}

\section{The FoundationChain Testbed}
\label{app:foundationchain}

\subsection{Targets.}

Each of the ten targets (Fig.~\ref{fig:foundationchain-grid}) probes a different aspect of crochet pattern writing. The `Alternating Swatch' and `Single Crochet/Half-Double Crochet/Double Crochet Swatch’ (\mbox{sc/hdc/dc}) are worked flat in rows of constant width and test fine-grained stitch recognition. The former uses double crochet stitches only, with each stitch placed in the space \emph{between} two stitches of the previous row. In the latter, rows alternate between the three stitch types. Also worked in rows, `Increasing' tests whether models can steadily increase the number of stitches at the end of each row to create a symmetrical tapered shape with even sides. `Circle' consists of concentric rounds of stitches and is the most basic test of working in the round. `Circle in Square', `Hexagon' and `Star' all test whether models can transition from a circle centre to edges of another shape. The `Granny Square’ is perhaps the archetypal crochet pattern: most crocheters learn it early, and the `granny' construction, whereby groups of stitches are worked into the spaces between groups in the previous round, is the basis for many more complex patterns and projects. The `Granny Triangle' adapts this approach for three sides. Finally, `Quarter Circle' is deliberately challenging: it begins in the same way as patterns worked in the round, then continues as if worked in rows.

\label{app:targets}
\begin{figure}[t]
  \centering
  \includegraphics[width=\columnwidth]{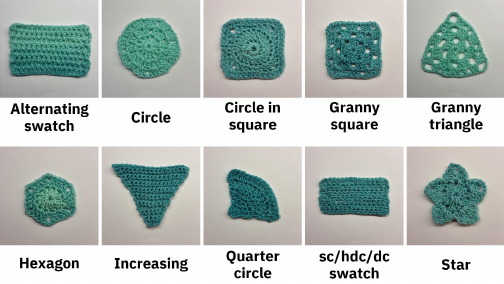}
  \caption{The FoundationChain targets.}
  \label{fig:targets-only}
\end{figure}

\subsection{Rules for Physical Execution}
\label{app:execution}

We follow each pattern’s instructions as literally as possible, without attempting to correct it. Where an instruction is malformed, ambiguous or incoherent, we apply a small set of rules. Every instruction that requires a rule is necessarily scored as an error under the relevant category of our taxonomy (\S\ref{sec:foundationchain}), typically Ambiguity, State Tracking or Count. All patterns are crocheted with Weight 5 Bulky/Chunky yarn and a 6mm crochet hook. We use a different colour to be able to easily distinguish different model outputs. Unlike a program that simply fails to compile, a pattern executed under these rules always yields a physical object that can be compared with the target. The rules are:

\paragraph{Insufficient stitches.} If a row or round calls for more stitches than there are positions available, we execute its instructions in order until no free positions remain, even partway through a repeat, then continue to the next row or round. 

\noindent\textit{Example:} Round 4 of GPT-5.6 Terra’s Granny Square requires eight ch-1 spaces, but Round 3 creates only four. Working in order, we complete a corner, a side, a second corner and a second side before the spaces run out, so Round 4 ends there, leaving two ch-2 spaces and two ch-1 spaces for Round 5.

\paragraph{Relative instructions.} Instructions such as `$x$ in the next $y$' or `skip $y$’ apply to the next $y$ stitches or spaces wherever they may be, and are ignored if no corresponding stitches or spaces exist. 

\noindent\textit{Example:} Row 1 of Kimi K3’s Alternating Swatch begins with a group of three dc and ends with a single dc, separated from the rest of the row by a chain-1 space. After turning, Row 2 begins at the final single dc of the last row. It calls for an FPdc (Front-Post double Crochet) around the first dc and then around each of the next 2 dc. As the next position is a ch-1 space, these two stitches are worked around the first two dc of the following group, passing over the space. The next instruction, `sk the ch-1 space’, then applies to the next ch-1 space, passing over the third dc of that group. Each such mismatch leaves multiple positions unused, so every row is shorter than the one before (\S\ref{par:geom}).

\paragraph{Contradictory conditions.} Where a pattern gives more than one condition for ending a repeat, row, round or pattern, and they cannot all be met at once, we stop as soon as any of them is met. 

\noindent\textit{Example:} Inkling’s Granny Triangle requires Row 3 to be repeated, `adding one ch-2 space and one 3-dc cluster on each side every row, until 10 rows are complete and 6 dc clusters are separated by ch-2 spaces across the row’. However, the second condition is met by Row 4, long before 10 rows or rounds are made, so we stop there.

\paragraph{Unmarked repeats.} Where a pattern gives an instruction to `repeat from *' or `repeat around’ without marking where the repeat begins, we repeat from the start of the round. Inkling does this in 5 outputs, and GPT-5.6 Terra in 2.

\paragraph{Undefined positions.} Where an instruction refers to a position that the pattern does not define and that cannot be identified in the piece (e.g., a border worked `in each space’ without defining what space or where), we omit that instruction. Multiple patterns attempt to make borders around the objects but do not specify where to begin the border in sufficient detail.

\subsection{Models and Sampling Settings}
\label{app:sampling}

We test four recent, widely-used proprietary and open-weight VLMs: Gemini 3.7 Flash (via the Gemini API; \citealp{gdm2026gemini37flash}), GPT-5.6 Terra  (via the OpenAI API; \citealp{openai2026gpt56}), Kimi K3 (via the Doubleword API; \citealp{moonshot2026kimik3}) and Inkling (via the Tinker API; \citealp{thinkingmachines2026inkling}). All outputs were generated between 31st of August and 3rd of September 2026. We generate one output per target from each model at temperature 0, with extended thinking disabled---except for Gemini 3.7 Flash, which does not allow this and for which we use thinking level `low' instead. The maximum output length is 2048 tokens (no outputs were truncated). The target image is supplied at every stage at a resolution of 1538$\times$1538 pixels. All other parameters are left at provider defaults. As API outputs are not guaranteed to be deterministic even at temperature 0, we will release every generated description, plan and pattern.

\subsection{Pattern Generation: Three-Stage Scaffold}
\label{app:scaffold}
Stages 1 (\textit{description}) and 2 (\textit{planning}) form a single conversation, so the same system prompt applies in both cases. Stage 3 (\textit{pattern writing}) is a separate call with its own system prompt, in which \texttt{{desc}} and \texttt{{plan}} are replaced with the model’s Stage 1 and Stage 2 outputs. 

\subsubsection{Stage 1: Description}
\begin{promptbox}[title={System}]
You are an expert crochet designer who can read a finished piece and reconstruct how it was made.
\end{promptbox}

\begin{promptbox}[title={User}]
Look at this crochet item. Reply in exactly this format, nothing else:
Pattern name: <a short, descriptive name for the pattern>
Description: <in no more than 3 sentences, cover: what kind of item it is (e.g., blanket, hat, scarf); its overall shape, size and proportions; the colour(s) and motifs; the texture and any other noticeable features>
\end{promptbox}

\subsubsection{Stage 2: Plan}
\begin{promptbox}[title={User}]
Based on this photo and the name and 
description you just gave, create a plan to 
recreate a pattern for this item --- there is 
no existing pattern, so you must work it out 
from the finished piece. Give short, clear 
answers with real numbers wherever you can:
1. What is the pattern type?
2. What is the overall shape?
3. Is it worked flat in rows, or in the round?
4. How does the shape change from beginning to end?
5. How do you start? Foundation chain or ring? How many stitches?
6. What are the main stitches used?
7. What does one typical row/round look like (stitches and count)?
8. Are any rows or sections repeated? Which rows, and how many times?
9. How many rows or rounds are there in total?
10. What hook size should be used?
11. What material is it made of?
\end{promptbox}

\subsubsection{Stage 3: Pattern Writing}
\begin{promptbox}[title={System}]
You are a professional crochet pattern writer. Examine the image and write crochet instructions in plain English prose, as printed in a published pattern. Write full natural-language sentences only.
\end{promptbox}
\begin{promptbox}[title={User}]
Using this photo and the description and plan below, write a complete published-style crochet pattern. Output only the pattern.
Description:
    {desc}
Plan:
    {plan}
\end{promptbox}

\section{Sample Outputs}
In this section, we provide complete outputs for the Hexagon target (Fig.~\ref{fig:foundationchain-grid}; \S\ref{par:geom}). We first give our own pattern for the target, which models are never shown, followed by reproductions of each model’s description, plan and pattern.

\subsection{Target}
\begin{promptbox}[title={Pattern: Hexagon}, listing and comment,
  comment={\centering\includegraphics[width=0.6\linewidth]{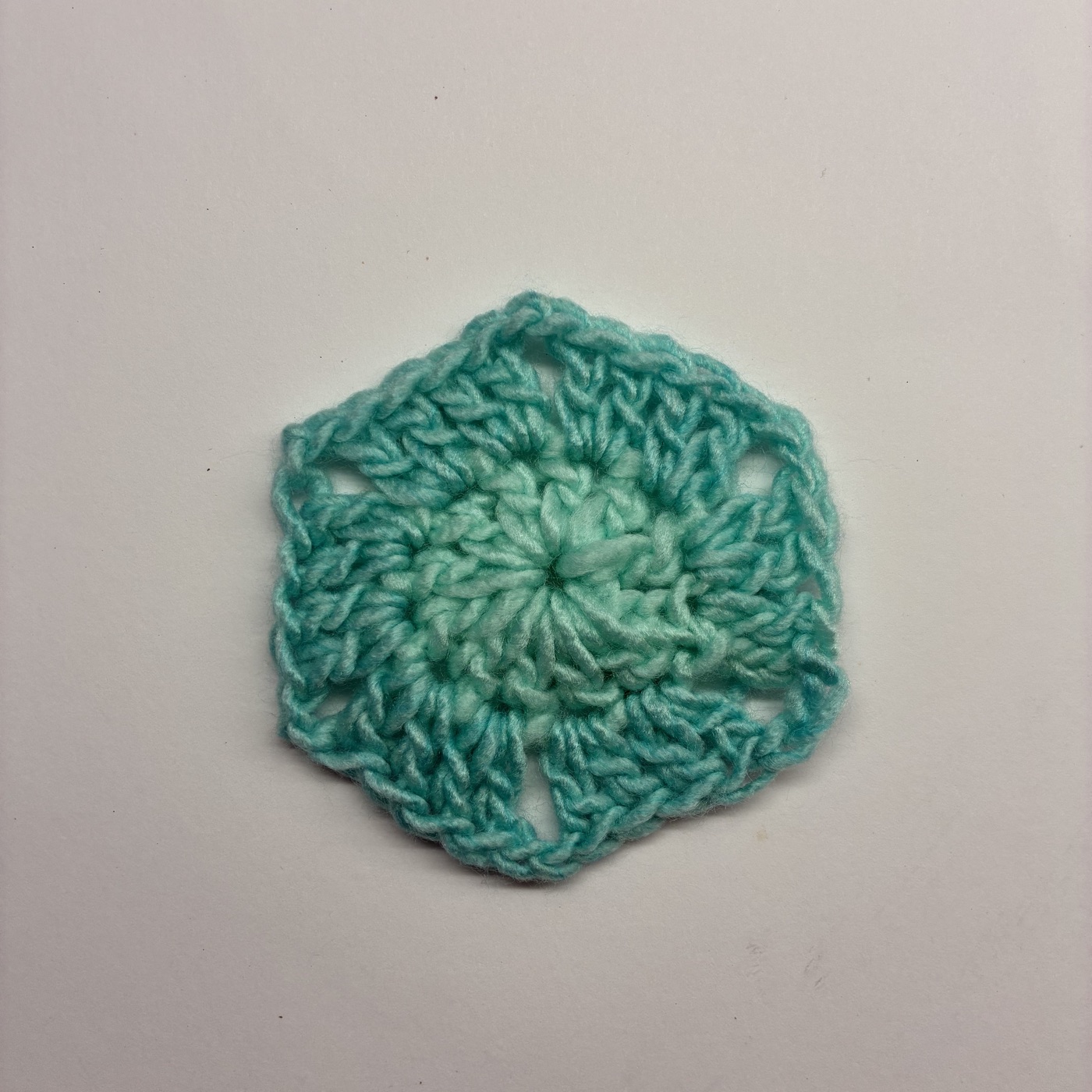}}]
Abbreviations: ch (chain), dc (double
crochet), slst (slip stitch).  

Begin with a magic ring, or ch 4 and join 
with a slst to make a ring.

Round 1: Ch 3 (counts as the first dc), then 
work 11 dc into the ring. Join with a slst 
to the top of the beginning ch-3. (12 dc)

Round 2: Slst into the next st, ch 3 (this 
counts as the first dc), then work 1 dc in
the same stitch. You will have two double 
crochets in the top of the first stitch.

Make another 2 dc in the next st (4dc total, 
two in each stitch), then ch 2 (this counts 
as the first corner). 

Make 2 more dc in each of the next 2 sts, 
then ch 2 (second corner).
Make another 2 dc in each of the next 2 sts, then ch 
2 (third corner).
Make 2 dc in each of the next 2 sts, then ch 
2 (fourth corner).
Make 2 dc in each of the next 2 sts, then ch 
2 (fifth corner).
Make 2 dc in each of the next 2 sts, then ch 
2 (sixth corner).
Join with a sl st to the top of the beginning 
ch-3 and fasten off. (24 dc total, or 6 groups of 4 dc 
separated by 6 ch-2 spaces)

\end{promptbox}
\subsection{Gemini 3.7 Flash}
\begin{promptbox}[title={Description}]
Pattern name: Solid Mini Hexagon Motif

Description: This is a small, flat hexagonal 
crochet motif worked in the round across two rounds. It features a gradient/variegated light mint and aqua-blue yarn with noticeable chain spaces creating the six distinct corners. The texture is mostly solid with standard double crochet stitches radiating outward from a closed center ring.    
\end{promptbox}
\begin{promptbox}[title={Plan}]
1. **Pattern type:** Geometric motif / Granny hexagon

2. **Overall shape:** Hexagon (6-sided regular polygon)

3. **Worked:** In the round (from center outwards), right side facing throughout

4. **Shape change:** Starts from a small circular center ring and expands into a 6-cornered hexagon via corner chain spaces and increases

5. **Start:** Magic ring (or ch 4, sl st to join); Round 1 begins with 12 double crochet (dc) stitches into the ring

6. **Main stitches used (US terms):** Magic ring, chain (ch), slip stitch (sl st), double crochet (dc)

7. **Typical round (Round 2):** Worked as 6 sides: `[1 dc in next 2 dc, (1 dc, ch 2, 1 dc) in corner space]` around (4 dc per side + 6 corner ch-2 spaces = 24 dc, 6 ch-2 spaces)

8. **Repeated sections:** Each round repeats the corner increase sequence `(1 dc, ch 2, 1 dc)` 6 times around the hexagon perimeter

9. **Total rounds:** 2 rounds

10. **Hook size:** 4.0 mm--4.5 mm (US G/6 or 7)

11. **Material:** Worsted/DK weight acrylic or cotton blend yarn in variegated aqua/mint green
\end{promptbox}
\begin{promptbox}[title={Pattern}, listing and comment,
  comment={\centering\includegraphics[width=0.6\linewidth]{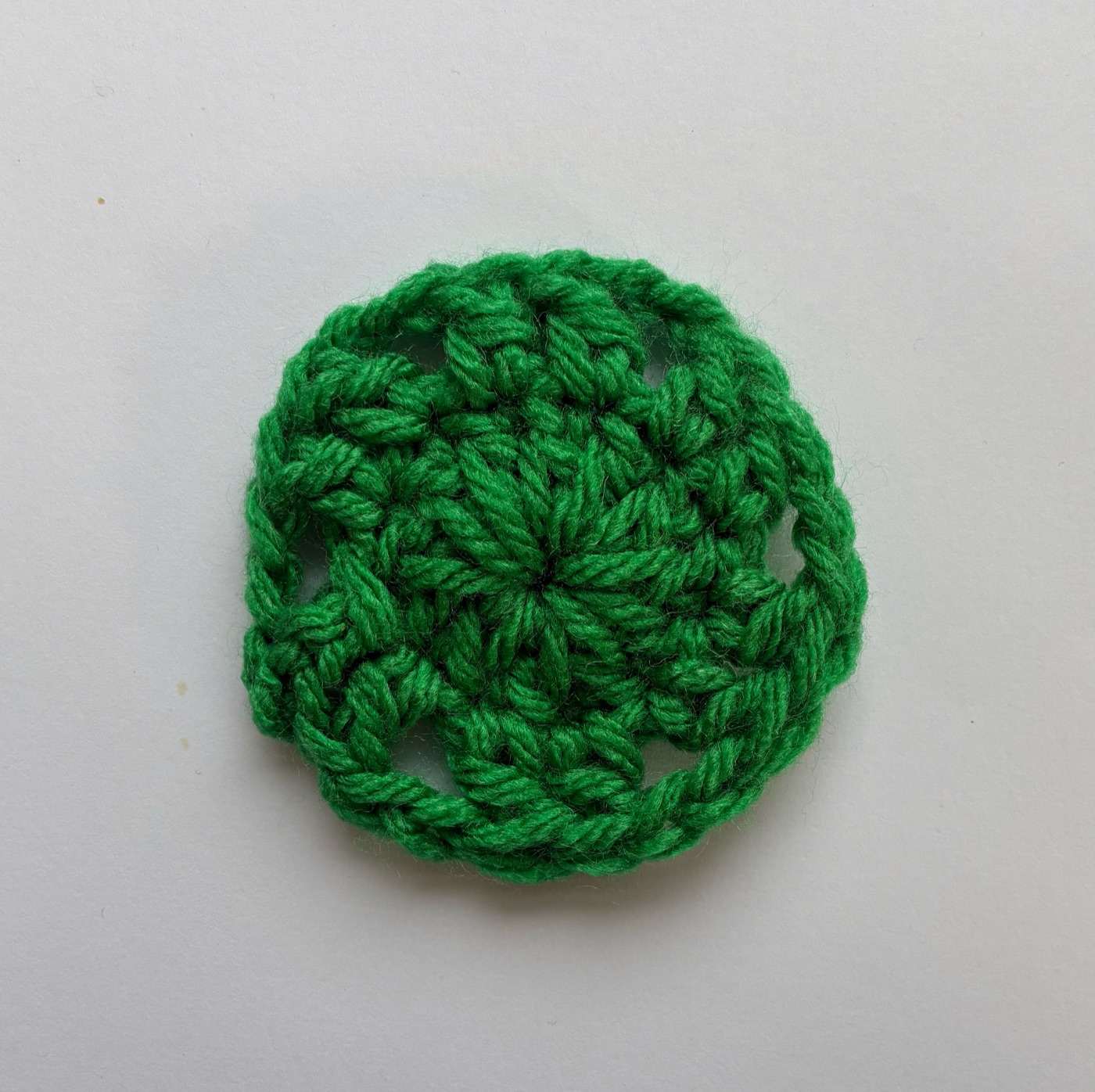}}]
**Solid Mini Hexagon Motif**

**Materials:**
* Worsted or DK weight yarn in a light aqua/mint variegated shade
* Crochet hook: Size US G/6 (4.0 mm) or 7 (4.5 mm), or size needed to obtain gauge
* Yarn needle
* Scissors

**Abbreviations (US Terms):**
* **ch:** chain
* **dc:** double crochet
* **sl st:** slip stitch
* **sp:** space
* **st(s):** stitch(es)

---

**Instructions:**

Begin with a magic ring (or ch 4 and sl st to the first ch to form a ring).

**Round 1:** Ch 3 (counts as first dc here and throughout), work 11 dc into the magic ring. Pull the tail firmly to close the center hole. Join with a sl st to the top of the beginning ch-3. (12 dc)

**Round 2:** Ch 3, dc in the next st, ch 2. *Dc in each of the next 2 sts, ch 2; repeat from * 4 more times around. Join with a sl st to the top of the beginning ch-3. (12 dc, 6 ch-2 corner spaces)

Fasten off and weave in all loose ends securely on the wrong side. Block lightly if desired to accentuate the 6 points of the hexagon.
\end{promptbox}

\subsection{GPT-5.6 Terra}
\begin{promptbox}[title={Description}]
Pattern name: Six-Petal Floral Granny Square  

Description: A small crochet granny-square motif, likely intended for joining into a blanket, cushion, or similar project. It is an approximately square aqua/turquoise piece with rounded corners, a raised six-petal flower-like centre, open corner spaces, and a textured cluster-stitch surface.
\end{promptbox}
\begin{promptbox}[title={Plan}]
1. **Pattern type:** Floral granny-square motif; a small modular square for a blanket, cushion, bag, or similar joined project.

2. **Overall shape:** Approximately square with softly rounded corners, about 8--10 cm / 3--4 in across in bulky yarn.

3. **Construction:** Worked in the round, with the right side facing throughout.

4. **Shape progression:** It begins as a compact circular flower centre, expands into a six-petal flower, then is squared off with four chain-space corners and straight cluster-stitch sides.

5. **Start:** Make a magic ring; work **12 double crochet (US dc)** into the ring and join with a slip stitch. A ch-4 ring could substitute, but the photo appears to have a closed adjustable centre.

6. **Main stitches:** Chain, slip stitch, US double crochet, likely **3-dc clusters/shells**, and chain-1 or chain-2 spaces; the petals may use puff/cluster stitches for their raised texture.

7. **Typical outer round:** In each corner space work **(3 dc, ch 2, 3 dc)**; along each side work **one 3-dc cluster/shell in each intervening space**, separated by ch 1. This gives 4 corners and roughly **2 side clusters per side** on the final round.

8. **Repeated sections:** The flower round repeats a petal unit **6 times** around the centre. The squaring round repeats the same side-and-corner unit **4 times**, once per side; there appears to be only one main square-forming round.

9. **Total rounds:** Approximately **4 rounds**: centre, petal foundation, raised petal round, and outer square round.

10. **Hook size:** About **5.5--6.5 mm (I/9--K/10 1/2)**, adjusted for a firm but not stiff fabric.

11. **Material:** Thick, softly spun aqua/turquoise yarn, likely bulky acrylic or an acrylic-wool blend.
\end{promptbox}
\begin{promptbox}[title={Pattern}, listing and comment,
  comment={\centering\includegraphics[width=0.6\linewidth]{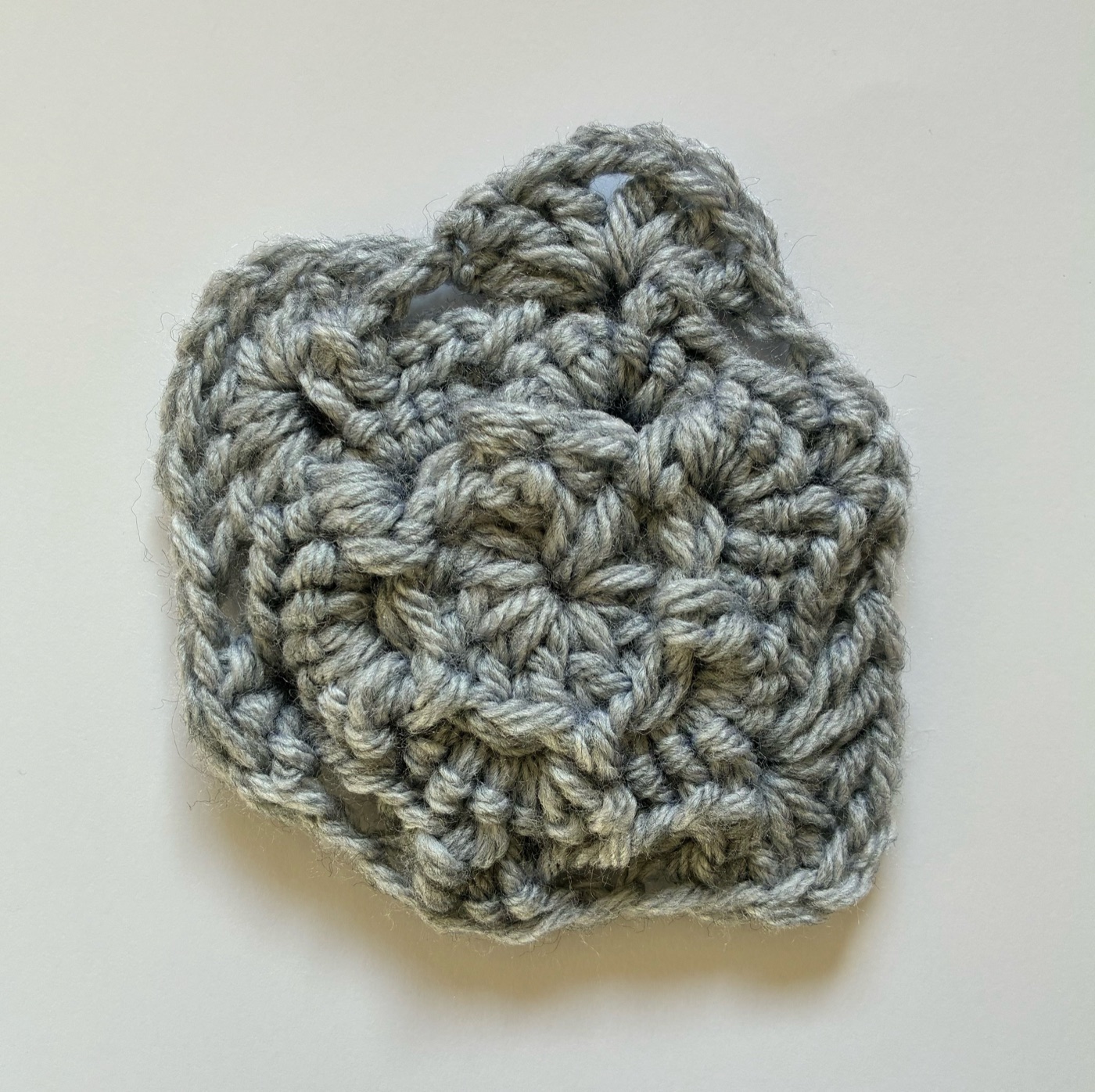}}]
# Six-Petal Floral Granny Square

This small textured motif is worked in the round with the right side facing throughout. Use bulky yarn and a 5.5--6.5 mm hook, or a hook size that gives a firm but flexible fabric.

**Abbreviations:** Ch = chain; dc = double crochet; sl st = slip stitch; st = stitch; sp = space.

**Special stitch---3-dc cluster:** Yarn over, insert hook in indicated space, yarn over and draw up a loop, yarn over and draw through 2 loops. Repeat twice more in the same space, then yarn over and draw through all 4 loops on hook.

**Round 1:** Make a magic ring. Ch 3, which counts as first dc, then work 11 dc into ring. Pull ring closed. Join with a sl st in top of beginning ch-3. You will have 12 dc.

**Round 2:** Ch 1. Work 1 dc in same st as joining. Ch 2, skip next st, work 1 dc in next st. Repeat from * around 5 more times, ending with ch 2. Join with a sl st in first dc. You will have 6 dc separated by 6 ch-2 spaces.

**Round 3:** Sl st into next ch-2 space. Ch 2. Work 1 3-dc cluster in same ch-2 space, ch 2, then work 1 more 3-dc cluster in same space. This forms the first petal. Work 2 3-dc clusters, separated by ch 2, in each remaining ch-2 space around. Join with a sl st in top of first cluster. You will have 6 petals.

**Round 4:** Sl st into the ch-2 space between the two clusters of the first petal. Ch 3, work 2 dc in same space, ch 2, then work 3 dc in same space to form the first corner.

Ch 1. In the ch-2 space between the clusters of the next petal, work 1 3-dc cluster. Ch 1. In the ch-2 space between the clusters of the following petal, work 1 3-dc cluster. Ch 1.

In the ch-2 space between the clusters of the next petal, work 3 dc, ch 2, and 3 dc. This forms the next corner.

Repeat the following twice more: Ch 1, work 1 3-dc cluster in each of the next 2 petal spaces, with ch 1 between clusters; then work 3 dc, ch 2, and 3 dc in the next petal space for a corner.

After the final corner, ch 1, work 1 3-dc cluster in each of the remaining 2 petal spaces, with ch 1 between clusters. Join with a sl st in top of beginning ch-3.

Fasten off and weave in ends. Block lightly to open the four corner spaces and shape the motif into a square.
\end{promptbox}

\subsection{Kimi K3}
\begin{promptbox}[title={Description}]
Pattern name: Classic Granny Square

Description: This is a small, flat crochet motif or coaster. It is shaped like a hexagon with six distinct corners and is made from a single solid color of light teal or mint green yarn. The texture features a central circle of stitches with openwork holes at the corners, characteristic of a traditional granny square construction.
\end{promptbox}
\begin{promptbox}[title={Plan}]
1. **Pattern type:** Granny square (hexagon variation) or motif-based coaster.

2. **Overall shape:** Hexagon with slightly rounded corners and scalloped edges.

3. **Construction:** Worked in the round, starting from the center.

4. **Shape progression:** Begins as a tight circle, expands into a hexagon through strategic corner increases in each round.

5. **Start:** Magic ring (or chain 4 joined to form a ring). Work 12 double crochet (DC) stitches into the ring (chain 3 counts as first DC).

6. **Main stitches:** Double crochet (DC), chain stitches (ch), and slip stitch (sl st) for joining.

7. **Typical round structure:** 

   - **Round 1:** 12 DC in ring (12 sts).

   - **Round 2:** *2 DC in next st, ch 1* around (24 sts, 12 ch-1 spaces).

   - **Round 3:** *3 DC in next ch-1 space, ch 2* around (6 clusters of 3 DC, 6 ch-2 corner spaces).

   - **Round 4:** *3 DC in next space, ch 2, 3 DC in same space (corner), ch 1, 3 DC in next space* around (establishes hexagon corners).

8. **Repetition:** Round 4 is repeated 2--3 times, working corner increases into each ch-2 corner space and side clusters into ch-1 spaces, until desired size is reached.

9. **Total rounds:** Approximately 4--5 rounds visible in the photo.

10. **Hook size:** 4.0 mm (G/6) to 5.0 mm (H/8), depending on yarn weight.

11. **Material:** Light worsted weight (DK or #3) acrylic or cotton yarn in mint/aqua.    
\end{promptbox}
\begin{promptbox}[title={Pattern}, listing and comment,
  comment={\centering\includegraphics[width=0.6\linewidth]{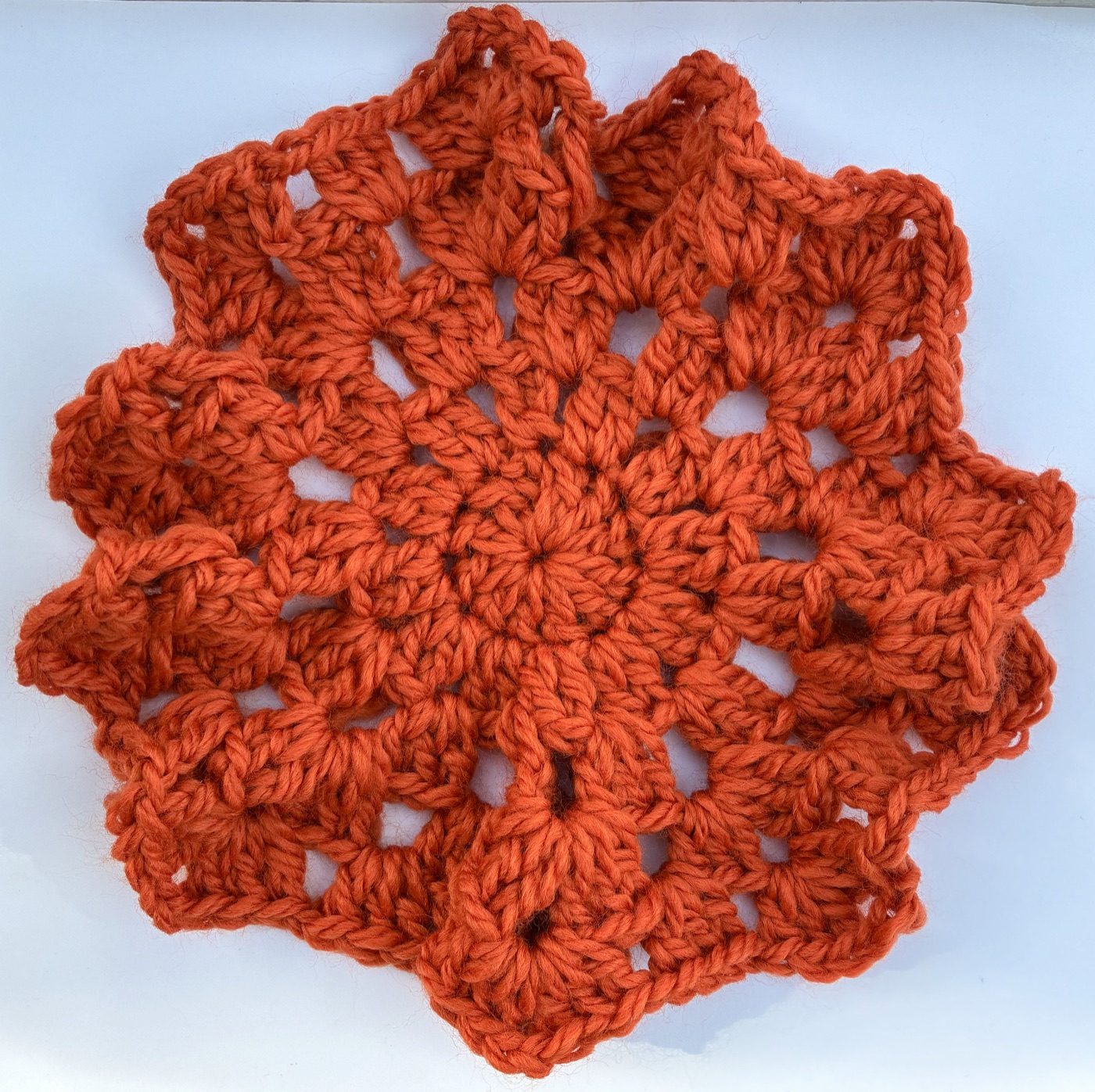}}]
**Classic Granny Square**



**Materials**

- Light worsted weight (DK or #3) acrylic or cotton yarn in mint/aqua

- Crochet hook size 4.0 mm (G/6) or 5.0 mm (H/8), depending on desired fabric density

- Yarn needle for weaving in ends

**Gauge**
Gauge is not critical for this motif. Finished size will vary with yarn weight and hook size.

**Finished Size**
Approximately 4--5 inches across, depending on materials used.

**Abbreviations**
- ch = chain
- dc = double crochet
- sl st = slip stitch
- st(s) = stitch(es)
- sp = space

**Pattern Notes**

This motif is worked in the round, beginning at the center. Do not turn unless otherwise indicated. Join each round with a slip stitch unless otherwise stated. Chain-3 at the beginning of a round counts as the first double crochet.

**Instructions**

**Round 1:** Make a magic ring (or ch 4 and join with sl st to form a ring). Ch 3 (counts as first dc), work 11 dc into ring. Join with sl st to top of beginning ch-3. (12 dc)

**Round 2:** Ch 3 (counts as first dc), dc in same st as join, ch 1. *Work 2 dc in next st, ch 1; repeat from * around. Join with sl st to top of beginning ch-3. (24 dc, 12 ch-1 sps)

**Round 3:** Sl st into first ch-1 sp. Ch 3 (counts as first dc), work 2 dc in same sp, ch 2. *Work 3 dc in next ch-1 sp, ch 2; repeat from * around. Join with sl st to top of beginning ch-3. (6 clusters of 3 dc, 6 ch-2 corner sps)

**Round 4:** Sl st into first ch-2 sp. Ch 3 (counts as first dc), work 2 dc in same sp, ch 2, work 3 dc in same sp (corner made), ch 1. *Work 3 dc in next ch-2 sp, ch 2, work 3 dc in same sp (corner made), ch 1; repeat from * around. Join with sl st to top of beginning ch-3. (6 corners, 6 ch-1 sps)

**Round 5:** Sl st into first ch-2 corner sp. Ch 3 (counts as first dc), work 2 dc in same sp, ch 2, work 3 dc in same sp (corner made), ch 1, work 3 dc in next ch-1 sp, ch 1. *Work 3 dc in next ch-2 corner sp, ch 2, work 3 dc in same sp (corner made), ch 1, work 3 dc in next ch-1 sp, ch 1; repeat from * around. Join with sl st to top of beginning ch-3.

**Finishing**

Fasten off, leaving a 6-inch tail. Weave in all ends securely with a yarn needle. Block lightly if desired to open the lace pattern and even the edges.
\end{promptbox}

\subsection{Inkling}
\begin{promptbox}[title={Description}]
Pattern name: Teal Hexagon Motif

Description: It is a small, flat hexagonal crochet motif in solid teal yarn with a textured, raised cluster-stitch surface. The piece measures roughly palm-sized with six straight sides and a small central opening. Its uniform color and open, lacy geometric pattern create a decorative, lightweight fabric.
\end{promptbox}
\begin{promptbox}[title={Plan}]
1. Pattern type: Hexagon crochet motif (e.g., coaster or afghan block).

2. Overall shape: Flat regular hexagon, about 4-5 in (10-12 cm) across.

3. Worked in the round.

4. Expands evenly from center; six sides stay flat with no cupping.

5. Magic ring; 6 clusters or 12 dc into ring.

6. Double crochet clusters/puffs with ch-1 or ch-2 spaces.

7. Round: (3 dc cluster, ch 2) in each space x 6 = 6 clusters, 6 ch-2 spaces.

8. Yes; cluster-in-space round repeats, adding 6 clusters each round.

9. About 5--6 rounds total.

10. 4.0 mm (G/6) hook.

11. Medium-weight cotton or acrylic yarn in solid teal.
\end{promptbox}
\begin{promptbox}[title={Pattern}, listing and comment,
  comment={\centering\includegraphics[width=0.6\linewidth]{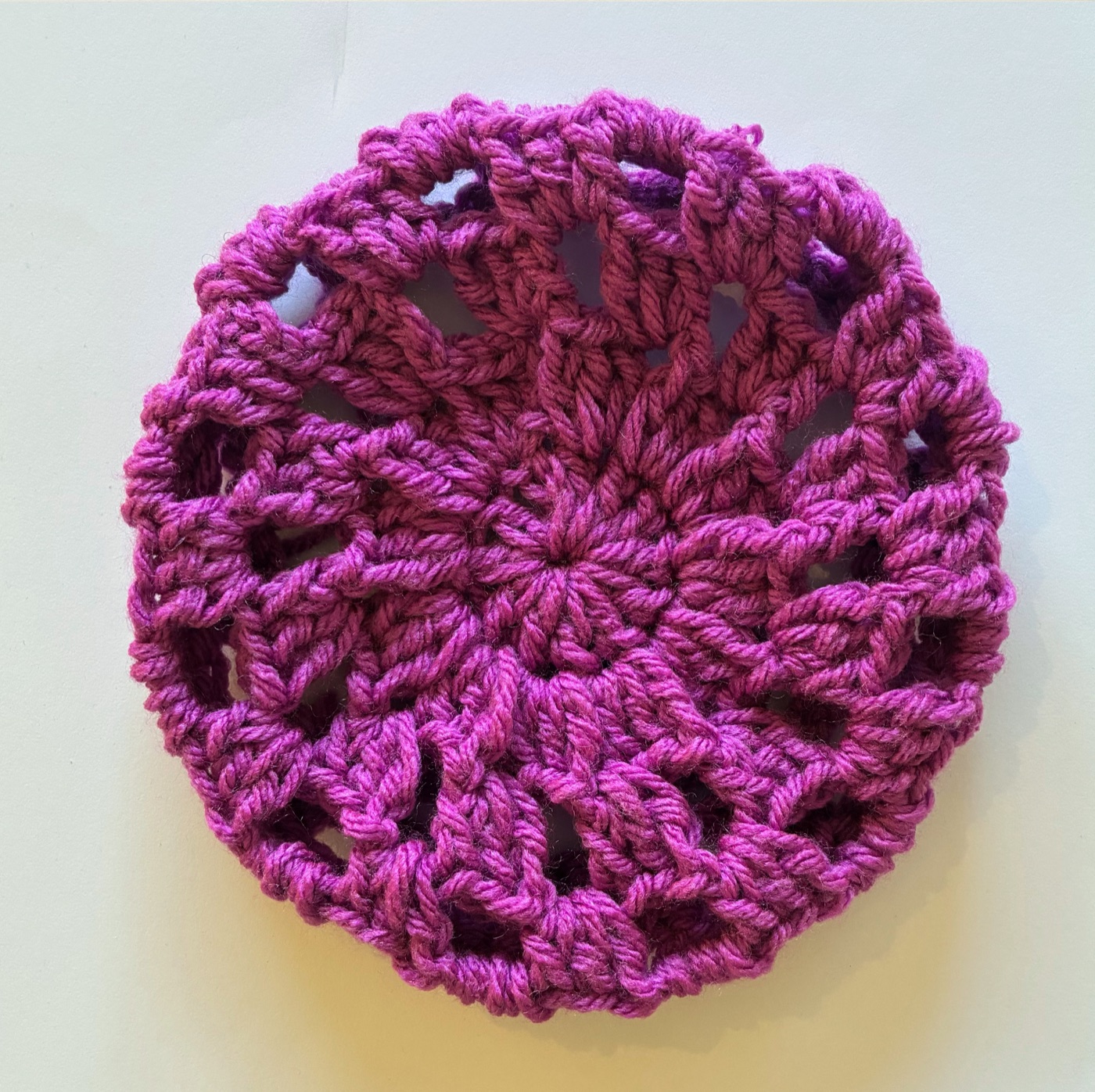}}]
Teal Hexagon Motif. 

Work in the round with a 4.0 mm hook and medium-weight teal yarn. 

Begin with a magic ring. 

Round 1: Ch 3, work 11 dc into the ring, sl st in the top of the beginning ch-3 to join; 12 dc. 

Round 2: Ch 3, work 2 dc in the same st, ch 2, then work 3 dc in each remaining st around with ch 2 between each cluster, sl st to join; 6 clusters and 6 ch-2 spaces. 

Round 3: Sl st into the next ch-2 space, ch 3, work 2 dc in the same space, ch 2, then work 3 dc in each ch-2 space around with ch 2 between clusters, sl st to join; 6 clusters. 

Round 4: Repeat Round 3. 

Round 5: Repeat Round 3. 

Round 6: Repeat Round 3. Fasten off and weave in ends. The finished motif measures about 4 to 5 inches across.

\end{promptbox}

\label{app:outputs}

\section{Definitions and Additional Experiments}
\subsection{Incorrect but Coherent Patterns}
\label{app:strongwrong}
We classify a pattern as incorrect but coherent if it produces the wrong object, but contains no Count, State tracking or Ambiguity errors and gives an object that lies flat. We include this last criterion because all targets are planar and no generated pattern deliberately sets out to make a piece that is not (13 of 40 patterns do so regardless; \S\ref{par:geom}). Of the 36 patterns with errors, 12 meet this definition:
\begin{itemize}[itemsep=0.5ex, topsep=0.5ex, parsep=0pt]
    \item \textit{Gemini 3.7 Flash (3):} Circle, Hexagon, \mbox{sc/hdc/dc} swatch
    \item \textit{GPT-5.6 Terra (5):} Alternating Swatch, Circle in Square, Increasing, Quarter circle, \mbox{sc/hdc/dc} swatch
    \item \textit{Kimi K3 (1):} Quarter circle
    \item \textit{Inkling (3):} Alternating Swatch, Circle in Square, Quarter circle
\end{itemize}
\begin{table*}[t]
\centering
\small
\setlength{\tabcolsep}{5pt}
\begin{tabular}{lcccccccccccc}
\hline
 & \multicolumn{4}{c}{Thinking off} & \multicolumn{4}{c}{Thinking low} & \multicolumn{4}{c}{Thinking medium} \\
\cline{2-5}\cline{6-9}\cline{10-13}
Model & $T{=}0$ & $0.7$ & $1.0$ & All & $T{=}0$ & $0.7$ & $1.0$ & All & $T{=}0$ & $0.7$ & $1.0$ & All \\
\hline
Gemini 3.7 Flash$^\dagger$ & --   & --   & --   & --   & 0.66 & 0.71 & 0.64 & \textbf{0.67} & --   & --   & --   & --   \\
GPT-5.6 Terra$^\ddagger$   & 1.74 & 1.86 & 1.86 & 1.84 & --   & --   & 2.03 & 2.03 & --   & --   & 1.95 & 1.95 \\
Kimi K3                    & 1.38 & 1.45 & 1.82 & 1.60 & 1.88 & 1.93 & 2.18 & 2.03 & 2.00 & 1.88 & 1.95 & 1.93 \\
Inkling                    & 1.84 & 1.77 & 1.90 & 1.83 & 2.06 & 2.22 & 1.99 & 2.10 & 2.02 & 2.03 & 1.97 & 2.00 \\
\hline
\end{tabular}
\caption{Mean absolute error of row/round counts when models are asked only to count the rows or rounds in each target photo. The targets have median 3.5, mean 5.4 and maximum 17 rows/rounds. \emph{All} pools the three temperatures. $^\dagger$Gemini 3.7 Flash cannot turn thinking off; low is its lowest setting. $^\ddagger$GPT-5.6 Terra accepts only $T{=}1.0$ with thinking enabled.}
\label{tab:rowcount}
\end{table*}
\subsection{Counting Rows and Rounds}
\label{app:rowcount}
What VLMs claim to see often depends on how the question is posed \citep{fan2026tinted}, and they perform worse when they must reason across modalities in a single step \citep{wang2026compose}. Since size errors are by far the most common type we encounter, we test the extent to which they stem from the scaffold, given that the number of rows or rounds is just one of eleven planning questions posed in a single prompt. We ask each model directly to count the rows or rounds in each of the 10 target images, using five variants of the same question:
\begin{enumerate}[itemsep=0.5ex, topsep=0.5ex, parsep=0pt]
\item \texttt{Please count the number of rows or rounds in this crochet object. Provide the number only.}
\item \texttt{How many rows or rounds were worked to make this crochet piece? Answer with a single number.}
\item \texttt{Count the rows (or rounds, if worked in the round) in the crocheted item in this photo. Reply with just the integer.}
\item \texttt{This is a photo of a crochet piece. How many rows or rounds does it have? Respond with the number only, no other text.}
\item \texttt{Looking at this crochet object, what is the total number of rows or rounds? Output only the number.}
\end{enumerate}

We run each prompt once at $T{=}0$ and three times at $T{=}0.7$ and at $T{=}1.0$, giving 35 answers per image. We report mean absolute error (MAE), averaged per image and then across images. Gemini 3.7 Flash cannot run with thinking disabled, so it always uses the low thinking budget. To ensure Gemini's performance cannot be wholly accounted for by the thinking budget afforded to it, we repeat the runs for other models with low and medium thinking. These thinking-on runs use the same three temperatures, except for GPT-5.6 Terra, which accepts only $T{=}1.0$ with thinking enabled. Gemini 3.7 Flash is significantly more accurate than every other model in every setting, and enabling thinking does not improve any model's results, consistent with findings that longer reasoning does not compensate for poor perception \citep{wu2026seeing}.

\end{document}